\documentclass{article}

\usepackage[nonatbib,preprint]{neurips_2026}
\usepackage[numbers,compress]{natbib}

\usepackage{amsmath}
\usepackage{amssymb}
\usepackage{amsthm}
\usepackage{booktabs}
\usepackage{microtype}
\usepackage{xcolor}
\usepackage{url}
\usepackage{multirow}
\usepackage{graphicx}
\usepackage{enumitem}

\theoremstyle{definition}
\newtheorem{definition}{Definition}
\theoremstyle{definition}
\newtheorem{hypothesis}{Hypothesis}

\newcommand{\SCACC}{0.97}
\newcommand{\CCACC}{0.28}
\newcommand{\CCFW}{0.63}
\newcommand{\QOACC}{0.06}
\newcommand{\CCPVAL}{$p{<}10^{-8}$}

\newcommand{\PHISCACC}{0.33}
\newcommand{\PHICCACC}{0.00}
\newcommand{\PHICCFW}{0.47}
\newcommand{\PHIQOACC}{0.08}
\newcommand{\PHICCPVAL}{$p{<}10^{-33}$}

\newcommand{\CTCCFW}{0.87}
\newcommand{\CTQOACC}{0.06}
\newcommand{\CTCTACC}{0.58}
\newcommand{\CTPVAL}{$p{<}10^{-27}$}

\newcommand{\CSSCACC}{0.82}
\newcommand{\CSCCACC}{0.10}
\newcommand{\CSCCFW}{0.76}
\newcommand{\CSQOACC}{0.09}
\newcommand{\CSPVAL}{$p{<}10^{-10}$}

\newcommand{\CSQWSCACC}{0.853}

\newcommand{\CSQWCCFW}{0.660}
\newcommand{\CSQWQOACC}{0.133}
\newcommand{\CSQWN}{150}

\newcommand{\GENNEXAMPLES}{300}
\newcommand{\GENNCORRECT}{147}
\newcommand{\GENNACC}{0.49}
\newcommand{\SGSCACC}{0.88}
\newcommand{\SGCCACC}{0.19}
\newcommand{\SGCCFW}{0.69}
\newcommand{\SGQOACC}{0.12}
\newcommand{\SGCCFWCILOW}{0.61}
\newcommand{\SGCCFWCIHIGH}{0.76}
\newcommand{\SGCCPVAL}{$p < 10^{-5}$}
\newcommand{\SGVSPREWRITTENPVAL}{0.240}

\newcommand{\ESPN}{100}
\newcommand{\ESPACCBASE}{0.02}
\newcommand{\ESPACCONE}{0.05}
\newcommand{\ESPACCTWO}{0.27}
\newcommand{\ESPACCTHREE}{0.53}
\newcommand{\ESPACCFOUR}{0.77}
\newcommand{\ESPACCFULL}{0.99}
\newcommand{\ESPEARLYRATIO}{0.05}

\newcommand{\FSN}{300}
\newcommand{\FSCA}{1.00}
\newcommand{\FSCB}{0.56}
\newcommand{\FSCC}{0.95}
\newcommand{\FSCD}{0.95}
\newcommand{\FSCBACC}{0.40}
\newcommand{\FSCDACC}{0.02}

\newcommand{\FSeightN}{150}
\newcommand{\FSeightCA}{1.000}
\newcommand{\FSeightCB}{0.647}
\newcommand{\FSeightCC}{0.987}
\newcommand{\FSeightCD}{0.920}
\newcommand{\FSeightCBacc}{0.353}
\newcommand{\FSeightCDacc}{0.047}

\newcommand{\SBSCACC}{0.99}
\newcommand{\SBCCACC}{0.00}
\newcommand{\SBCCFW}{1.00}
\newcommand{\SBQOACC}{0.05}
\newcommand{\SBCTACC}{0.99}
\newcommand{\SBN}{200}

\newcommand{\MISTSCACC}{1.00}
\newcommand{\MISTCCACC}{0.02}
\newcommand{\MISTCCFW}{0.98}
\newcommand{\MISTQOACC}{0.05}
\newcommand{\MISTN}{500}

\newcommand{\PHIFOURSCACC}{0.850}
\newcommand{\PHIFOURCCACC}{0.390}
\newcommand{\PHIFOURCCFW}{0.300}
\newcommand{\PHIFOURQOACC}{0.000}
\newcommand{\PHIFOURCCPVAL}{$p{<}10^{-8}$}
\newcommand{\PHIFOURN}{100}

\newcommand{\QWTBSCACC}{0.960}
\newcommand{\QWTBCCACC}{0.940}
\newcommand{\QWTBCCFW}{0.010}
\newcommand{\QWTBQOACC}{0.300}
\newcommand{\QWTBN}{100}

\newcommand{\QTHREENSBASE}{0.997}
\newcommand{\QTHREENSMID}{0.983}
\newcommand{\QTHREENSPRE}{0.997}
\newcommand{\QTHREENSSUF}{0.970}
\newcommand{\QTHREENSQO}{0.010}
\newcommand{\QTHREESCACC}{1.000}
\newcommand{\QTHREECCACC}{0.270}
\newcommand{\QTHREECCFW}{0.730}
\newcommand{\QTHREEQOACC}{0.000}
\newcommand{\QTHREEN}{100}

\newcommand{\QTHREEFOURN}{200}
\newcommand{\QTHREEFOURSCACC}{1.000}
\newcommand{\QTHREEFOURCCACC}{0.520}
\newcommand{\QTHREEFOURCCFW}{0.480}
\newcommand{\QTHREEFOURQOACC}{0.075}
\newcommand{\QTHREEFOURPVAL}{$p{<}10^{-10}$}

\newcommand{\MATHORIGbase}{0.690}
\newcommand{\MATHORIGprefix}{0.720}
\newcommand{\MATHORIGmiddle}{0.660}
\newcommand{\MATHORIGsuffix}{0.060}
\newcommand{\MATHORIGsuffixFW}{0.087}
\newcommand{\MATHSTRIPbase}{0.540}
\newcommand{\MATHSTRIPprefix}{0.560}
\newcommand{\MATHSTRIPmiddle}{0.390}
\newcommand{\MATHSTRIPsuffix}{0.510}
\newcommand{\MATHSTRIPsuffixFW}{0.944}
\newcommand{\MATHQO}{0.010}
\newcommand{\MATHN}{100}

\newcommand{\DSSCACC}{0.990}
\newcommand{\DSCCACC}{0.020}
\newcommand{\DSFW}{0.980}

\newcommand{\QWSBVONEFW}{0.590}

\newcommand{\QBVONESCACC}{1.000}
\newcommand{\QBVONECCACC}{0.930}
\newcommand{\QBVONEFW}{0.060}
\newcommand{\QBVONEQOACC}{0.270}
\newcommand{\QBVONEN}{100}
\newcommand{\QBVONEPVAL}{$p{=}0.384$}

\newcommand{\QBASEVONESCACC}{0.990}
\newcommand{\QBASEVONECCACC}{0.090}
\newcommand{\QBASEVONEFW}{0.900}
\newcommand{\QBASEVONEQOACC}{0.000}
\newcommand{\QBASEVONEN}{100}
\newcommand{\QBASEVONEPVAL}{$p{<}10^{-10}$}

\newcommand{\PCTXN}{100}
\newcommand{\PCTXSTDACC}{0.930}
\newcommand{\PCTXSTDFW}{0.000}
\newcommand{\PCTXCACC}{0.940}
\newcommand{\PCTXCFW}{0.020}
\newcommand{\PCTXSUFFW}{0.590}
\newcommand{\PCTXSUFACC}{0.410}
\newcommand{\PCTXPOSPVAL}{$p{<}10^{-10}$}
\newcommand{\PCTXFWRED}{0.570}
\newcommand{\DSQOACC}{0.010}
\newcommand{\DSPVAL}{$p{<}10^{-12}$}
\newcommand{\DSN}{100}
\newcommand{\CSAFULL}{0.82}
\newcommand{\CSANEUTRAL}{0.32}
\newcommand{\CSACORRLAST}{0.29}
\newcommand{\CSACORRPEN}{0.83}
\newcommand{\CSANOCONC}{0.35}
\newcommand{\CSAGAP}{54.0}
\newcommand{\CSAGAPPVAL}{$p{<}10^{-22}$}

\newcommand{\GENPBN}{200}
\newcommand{\GENPBGENACC}{0.56}
\newcommand{\GENPBACCBASE}{0.015}
\newcommand{\GENPBACCONE}{0.02}
\newcommand{\GENPBACCTWO}{0.14}
\newcommand{\GENPBACCTHREE}{0.27}
\newcommand{\GENPBACCFOUR}{0.24}
\newcommand{\GENPBECR}{0.035}

\newcommand{\GENPBsBN}{200}
\newcommand{\GENPBsBGENACC}{0.49}
\newcommand{\GENPBsBACCBASE}{0.015}
\newcommand{\GENPBsBACCONE}{0.02}
\newcommand{\GENPBsBACCTWO}{0.08}
\newcommand{\GENPBsBACCTHREE}{0.14}
\newcommand{\GENPBsBACCFOUR}{0.16}
\newcommand{\GENPBsBECR}{0.041}

\newcommand{\MFCHAIN}{0.95}
\newcommand{\MFFILLERONE}{0.75}
\newcommand{\MFFILLERTWO}{0.61}
\newcommand{\MFFILLERTHREE}{0.70}
\newcommand{\MFQO}{0.06}
\newcommand{\MFFILLERAVG}{0.69}
\newcommand{\MFSPREAD}{0.14}
\newcommand{\MFN}{500}
\newcommand{\MFPVAL}{= 0.001}

\newcommand{\BIDN}{60}
\newcommand{\BIDSCACC}{0.567}
\newcommand{\BIDCCACC}{0.000}
\newcommand{\BIDCCFW}{0.617}
\newcommand{\BIDQOACC}{0.083}
\newcommand{\BIDCTACC}{0.167}
\newcommand{\BIDDELTA}{+40.0}

\newcommand{\CPFWPRE}{0.05}
\newcommand{\CPFWMID}{0.01}
\newcommand{\CPFWSUF}{0.39}
\newcommand{\CPFWREPLW}{0.32}   
\newcommand{\CPFWNEUTRAL}{0.00}   

\newcommand{\CPN}{500}

\newcommand{\CPSUFPREPVAL}{< 0.001}
\newcommand{\CPSCPRE}{0.15}
\newcommand{\CPSCMID}{0.53}
\newcommand{\CPSCSUF}{0.94}

\title{The Last Word Often Wins: A Format Confound in Chain-of-Thought Corruption Studies}

\author{
  Gabriel Garcia \\
  Independent Researcher \\
  \texttt{gpgabriel25@gmail.com}
}

\begin{document}

\maketitle

\begin{abstract}
Corruption studies, the standard tool for evaluating chain-of-thought (CoT)
faithfulness, infer which steps are ``computationally important'' from accuracy
loss when steps are corrupted. We show that when benchmark chains end with an
explicit terminal answer line, as in GSM8K and MATH, these tests largely
measure \emph{answer placement} rather than where intermediate computation is
carried out.

Using matched GSM8K examples, removing only the final answer statement while
preserving all reasoning collapses suffix sensitivity by about $19\times$ for
Qwen~2.5-3B ($N{=}300$, $p{=}0.022$). Conflicting-answer prompts, which contain
correct reasoning but a wrong explicit final answer, drive accuracy to zero or
near-zero at 7B across five open-weight model families; wrong-answer following is
strong at 3B--7B and attenuates sharply at larger scales. Replications on MATH,
within-stable comparisons at 7B, and suffix-free chains show the same pattern in
different guises: corruption sensitivity tracks the location of explicit answer
text, not a fixed computational depth in the reasoning.

Generation-time probes indicate that final answers are rarely early-determined
during generation (${<}5\%$ early commitment), yet consumption-time behavior
systematically follows explicit answer text. The confound is therefore largely a
readout effect when the chain is consumed. We propose a three-prerequisite protocol
(question-only control, format characterization, and an all-position sweep) as a
practical minimum for future corruption-based faithfulness studies.
\end{abstract}

\section{Introduction}
\label{sec:intro}

Chain-of-thought (CoT) prompting elicits step-by-step reasoning from large
language models~\citep{wei2022}. Corruption studies, replacing specific steps
with errors and measuring the accuracy drop, are the primary empirical tool
for evaluating whether these steps are computationally
meaningful~\citep{lanham2023,turpin2023,pfau2024}. Process reward
models~\citep{lightman2023} and faithfulness evaluations~\citep{turpin2023}
depend on these studies to assign credit to individual reasoning steps.

We identify a systematic confound: for chains with explicit terminal answer
statements (the dominant format in benchmarks like GSM8K and MATH), corruption
studies detect where the answer text appears, not where computation occurs.
Stripping only the answer statement from GSM8K chains while preserving all
reasoning collapses suffix sensitivity nearly $20\times$ at 3B
(Figure~\ref{fig:reversal}). Conflicting-answer experiments confirm the
mechanism: models systematically follow wrong terminal answers over correct
intermediate computation, with the effect strongest at 3B--7B and attenuating
at larger scales (followed-wrong spans \CCFW--\SBCCFW{} at 3B--7B; Phi-4-14B
FW\,$=$\,\PHIFOURCCFW; Qwen~2.5-32B FW\,$\approx$\,\QWTBCCFW). The 7B format-ablation counterpart
shows $9.3\times$ within-stable attenuation ($N{=}76$, $p{=}7.8{\times}10^{-3}$),
replicated in Qwen3-8B ($N{=}299$, $p{=}0.004$). On MATH (DeepSeek-R1-7B) the same
cross-format pattern yields $10.9\times$ suffix-survival recovery; on suffix-free
Hard-v3 chains the protocol instead marks the prefix as load-bearing
($\Delta{=}{-}0.77$, $p{<}10^{-12}$). Generation-time probes show early answer
commitment ${<}\,5\%$ while consumption-time outputs still track explicit answer
text; format-determination persists through 14B ($8.5\times$ ratio, $p{=}0.001$)
before both override and format effects shrink toward zero at 32B. We distill
these findings into a three-prerequisite protocol (question-only control, format
characterization, and an all-position sweep) that should be standard for any
corruption-based faithfulness evaluation.

\textbf{Contributions.} (i)~We identify and experimentally isolate a previously
undocumented format confound in CoT corruption studies. (ii)~We provide causal
evidence from four complementary designs: within-dataset format ablation, a
$2{\times}2$ reasoning-by-answer-line factorial, conflicting-answer tests
across five model families, and answer-placement controls. (iii)~We
demonstrate that a published-style corruption protocol produces qualitatively
different positional conclusions under format control. (iv)~Generation-time
probes show that answers are not early-determined, bounding the interpretation
to answer-text readout dominance at consumption time rather than early answer
commitment during generation.

\textbf{Interpretation.} The evidence supports answer-text readout dominance:
models can compute through intermediate steps during generation, but at readout
their behavioral output preferentially tracks explicit answer text rather than
re-deriving the conclusion from reasoning. Whether this reflects a general
rationalization pattern or a narrower answer-presentation heuristic shaped by
instruction tuning remains an open question (\S\ref{sec:limitations}).

\textbf{Scope.} All primary claims in this paper are scoped to chain-of-thought
formats that end with an explicit terminal answer line (``The answer is~$X$.''),
the dominant format in standard reasoning benchmarks, and to the 3B--32B
open-weight scale range we test. Generalizability to heterogeneous, naturally
occurring CoT \emph{without} explicit answer-line endings is not directly tested;
where chains lack such suffixes, the bidirectional ablation
(\S\ref{sec:bidirectional}) shows that the prefix becomes load-bearing instead,
consistent with sensitivity tracking answer-text location rather than position
per se. See~\S\ref{sec:limitations} for the full scope discussion.

\begin{figure}[!htbp]
  \centering
  \small
  \begin{minipage}[t]{0.47\linewidth}
    \centering
    \textbf{GSM8K-v1 (diagnostic illustration, $N{=}100$)}\\[4pt]
    \begin{tabular*}{\linewidth}{@{\extracolsep{\fill}}lc@{}}
      \toprule
      Condition & Acc \\
      \midrule
      Baseline (chain) & .970 \\
      \colorbox{red!25}{Suffix corrupt} & \colorbox{red!25}{.210} \\
      \bottomrule
      \multicolumn{2}{@{}p{\linewidth}@{}}{\scriptsize Qwen~2.5-3B; suffix-only slice of Table~\ref{tab:main-results}. Not the matched-$N{=}300$ accuracy pair used for the headline $\Delta_{\mathrm{suffix}}$ (see caption).}
    \end{tabular*}
  \end{minipage}\hfill
  \begin{minipage}[t]{0.47\linewidth}
    \centering
    \textbf{GSM8K-stripped-v1 (matched $N{=}300$)}\\[4pt]
    \begin{tabular*}{\linewidth}{@{\extracolsep{\fill}}lcc@{}}
      \toprule
      Condition & Acc & $\Delta_{\mathrm{suf}}$ \\
      \midrule
      Baseline (chain) & .960 & --- \\
      Suffix corrupt & .920 & $-$.040 \\
      \bottomrule
      \multicolumn{3}{@{}l@{}}{\scriptsize Qwen~2.5-3B (Table~\ref{tab:ablation}).}
    \end{tabular*}
  \end{minipage}
  \caption{\textbf{Within-dataset format ablation (Qwen~2.5-3B).} The headline
  ${\approx}19{\times}$ reduction compares matched-$N{=}300$ \emph{suffix} $\Delta$
  on the standard-format branch ($\Delta_{\mathrm{suffix}}=-0.760$) to
  GSM8K-stripped-v1 ($\Delta_{\mathrm{suffix}}=-0.040$, $p{=}0.022$, sign test;
  Table~\ref{tab:ablation}). The \emph{right} mini-table shows the corresponding
  matched-$N{=}300$ accuracies for the stripped format. The \emph{left} mini-table
  is a \emph{diagnostic illustration} only: suffix-only accuracies from the
  $N{=}100$ GSM8K-v1 slice (Table~\ref{tab:main-results}) that make the qualitative
  pattern easy to read; it does \emph{not} print the matched-$N{=}300$
  standard-format accuracy pair (those full grids are in Table~\ref{tab:ablation}).
  Full prefix/middle/suffix breakdowns: Tables~\ref{tab:main-results}
  and~\ref{tab:ablation}. Cross-dataset reversal (Hard-v3 vs.\ GSM8K-v1):
  Table~\ref{tab:main-results}.}
  \label{fig:reversal}
\end{figure}

\section{Related Work}
\label{sec:related}

\paragraph{Chain-of-thought faithfulness and causal probes.}
\citet{turpin2023} showed that CoT can be unfaithful when models are biased by
few-shot demonstrations: the generated chain does not always match the factors
actually driving the prediction. \citet{ye2022} showed that post-hoc
explanations from NLP models are unreliable indicators of underlying
computation. \citet{lanham2023} showed that truncating or corrupting CoT steps
can hurt accuracy, suggesting at least some chain content plays a causal role.
\citet{pfau2024} explored hidden computation in transformer language models,
showing that filler tokens can serve as implicit computation steps.
\citet{baker2025} recently proposed monitoring reasoning faithfulness
by comparing the chain-of-thought to a model's internal activations,
providing a complementary perspective to behavioral corruption studies.
None of these works control for the confound we identify:
that positional corruption sensitivity may reflect answer placement in the
chain format rather than computational structure. To our knowledge, no
prior work has demonstrated that the same corruption protocol yields
opposite positional conclusions on chains with different answer-placement
formats, nor has any work proposed a format ablation to isolate the
mechanism.

\paragraph{CoT prompting and emergent reasoning.}
\citet{wei2022} introduced CoT prompting; \citet{wang2022} extended it with
self-consistency decoding; \citet{kojima2022} demonstrated zero-shot CoT.
\citet{madaan2022} decomposed chain contributions (symbols, patterns, text)
but did not characterize which positions carry the explicit answer.
These works establish that chain content correlates with accuracy but leave
causal attribution open.

\paragraph{Theoretical capacity.}
\citet{merrill2023} proved that CoT extends transformer capacity beyond
$\mathrm{TC}^0$, but this is a bound on \emph{existence}, not on whether
empirical chains use that capacity at the probed positions.
\citet{saparov2023} showed systematic errors in multi-step inference,
suggesting that how the written chain relates to underlying computation is complex.

\paragraph{Process supervision.}
\citet{lightman2023} and \citet{uesato2022} assign step-level credit
via process reward models, implicitly assuming steps are causally meaningful
at their textual positions. \citet{goyal2024} showed that learnable pause
tokens provide computation without interpretable steps. Our finding
that causal sensitivity tracks answer placement suggests process reward
models may disproportionately reward answer expression over reasoning.

\paragraph{Sycophancy and instruction following.}
A related but distinct phenomenon is sycophancy, where models adapt
outputs to match perceived user preferences regardless of
correctness~\citep{perez2023sycophancy,sharma2023sycophancy}.
\citet{sharma2023sycophancy} showed that RLHF training amplifies this
tendency, causing models to endorse user-expressed views even when
incorrect. Unlike sycophancy studies, which investigate social-desirability
bias in open-ended dialogue, our work isolates answer-text \emph{placement}
as a methodological confound in chain-of-thought corruption
experiments, showing that sensitivity to corrupted reasoning reflects
consumption of explicit answer tokens embedded in the chain format,
rather than deference to an experimenter's social preference.

\paragraph{Gap.} No prior work has (i)~demonstrated that chain format
determines positional sensitivity in corruption studies, (ii)~proposed
format-awareness controls for corruption-based causal analysis,
(iii)~shown that a format ablation can shift the corruption
sensitivity pattern, isolating answer placement as the dominant
causal factor, or (iv)~provided a direct conflicting-answer causal test
showing that models prioritize explicit answer text over their own
correct computation at \emph{consumption time}.

\section{Problem Formulation}
\label{sec:formulation}

\subsection{Answer-Text Readout Dominance}

We define the central question this paper investigates. Let $q$ denote a
question and $c = (s_1, \ldots, s_K)$ a chain of $K$ reasoning steps. A model
$f$ maps $(q, c) \to \hat{a}$ to produce a final answer.

\begin{hypothesis}[Answer-Text Readout Dominance]
\label{hyp:rationalization}
  At \emph{consumption time}, when a model reads a completed chain, the
  model's final answer $\hat{a}$ is primarily determined by \emph{explicit
  answer text} present in the chain (e.g., a terminal ``the answer is $X$''
  statement), rather than by the intermediate computation encoded in steps
  $s_1, \ldots, s_{K-1}$. This describes a behavioral regularity, the
  model's output systematically tracks the explicit answer, without
  requiring a specific cognitive mechanism (e.g., instruction-following,
  format-completion heuristics, or recency-weighted readout are all
  compatible with this regularity). Letting $a_{\mathrm{exp}}$ denote the
  explicit answer text in $c$:
  \[
    \hat{a} \;\approx\; f_{\mathrm{readout}}(q,\, a_{\mathrm{exp}}),
  \]
  with intermediate reasoning steps playing a reduced causal role at
  readout, even when they encode the correct answer through genuine
  step-by-step computation.
\end{hypothesis}

If answer-text readout dominance holds, a key behavioral
prediction follows: models will be sensitive to corruptions wherever the chain
\emph{expresses} the answer text, not wherever intermediate computation
\emph{occurs}. This is consistent with a model that genuinely computes
through intermediate steps during generation, yet at readout preferentially
tracks the explicit answer signal rather than re-deriving the conclusion from
embedded reasoning. Our experiments test this prediction directly.

\subsection{Notation}

Accuracy on a task slice $\mathcal{D} = \{(q_i, c_i, a_i^*)\}_{i=1}^N$ is
\[
  \mathrm{acc}(f, \mathcal{D}) \;=\;
  \frac{1}{N}\sum_{i=1}^N \mathbf{1}\bigl[f(q_i, c_i) = a_i^*\bigr].
\]

\paragraph{Corruption operator.}
For a position subset $P \subseteq [K]$, define
$c^P = \mathrm{corrupt}(c, P)$ as the chain with steps at positions $P$
replaced by semantically incorrect but syntactically valid alternatives (e.g.,
wrong arithmetic, incorrect logical conclusions). We study three canonical
position subsets:
\begin{itemize}
  \item \textit{Prefix} ($P_\mathrm{pre}$): steps in the first third of the chain.
  \item \textit{Middle} ($P_\mathrm{mid}$): steps in the middle third.
  \item \textit{Suffix} ($P_\mathrm{suf}$): steps in the final third.
\end{itemize}

\paragraph{Question-only condition.}
Let $f_\emptyset(q)$ denote the model's answer given \emph{only} the question,
with no chain provided. Accuracy under the question-only condition is
\[
  \mathrm{acc}_\emptyset(\mathcal{D}) \;=\;
  \frac{1}{N}\sum_{i=1}^N \mathbf{1}\bigl[f_\emptyset(q_i) = a_i^*\bigr].
\]

\subsection{The Question-Solvability Confound}

\begin{definition}[Question-solvability confound]
  A corruption study on slice $\mathcal{D}$ suffers the
  \emph{question-solvability confound} when
  \(
    \mathrm{acc}(f,\mathcal{D}) - \mathrm{acc}_\emptyset(\mathcal{D}) \leq \epsilon
  \)
  for a small threshold $\epsilon \geq 0$.
\end{definition}

When the confound holds, observed robustness of
$\mathrm{acc}(f(\cdot, c^P), \mathcal{D})$ to corruption cannot be attributed
to chain causal structure: the model answers primarily from the question, not
the chain.

\subsection{Control Requirement and Interpretability}

\begin{definition}[Control requirement]
\label{def:2}
  A corruption study produces interpretable positional evidence only if
  \[
    \mathrm{acc}(f, \mathcal{D}) - \mathrm{acc}_\emptyset(\mathcal{D})
    \;>\; \epsilon_\mathrm{min},
  \]
  where $\epsilon_\mathrm{min} > 0$ is a meaningful minimum gap established
  by a statistical test rejecting
  $H_0\colon \mathrm{acc}(f) = \mathrm{acc}_\emptyset$.
\end{definition}

\begin{definition}[Causally load-bearing position]
  Under the control requirement, position $P$ is \emph{causally load-bearing}
  for $\mathcal{D}$ if
  \[
    \mathrm{acc}(f, \mathcal{D}) - \mathrm{acc}(f(\cdot, c^P), \mathcal{D})
    \;>\; 0
  \]
  with sufficient statistical confidence on a matched paired test.
\end{definition}

This formalization requires \emph{two} sequential tests: (i)~the chain-gap
test (control requirement), then (ii)~the positional-drop test (causal
load-bearing test). A study reporting only (ii) without (i) cannot support
positional causal claims.

\subsection{The Answer-Placement Prediction}

A key testable prediction of Hypothesis~\ref{hyp:rationalization} concerns
how models behave in corruption studies. If a model is answer-text-prioritizing
at readout, it will be sensitive to corruptions at positions where the
answer is \emph{expressed}, regardless of where computation logically occurs.
This yields:

\begin{hypothesis}[Answer-placement prediction]
\label{hyp:format}
    The set of positions identified as ``causally load-bearing'' by a
    corruption study is determined by where the chain format explicitly encodes
    the final answer, not by where genuine intermediate computation occurs.
\end{hypothesis}

This hypothesis generates two testable predictions: (a)~chains that
place the answer in the prefix should show prefix sensitivity;
chains that place the answer in the suffix should show suffix
sensitivity; and (b)~modifying the format (e.g., removing
explicit answer statements) should shift the sensitivity pattern.

A third, stronger prediction follows directly from
Hypothesis~\ref{hyp:rationalization}:
\begin{hypothesis}[Direct conflict prediction]
\label{hyp:conflict}
    When correct reasoning steps conclude with a conflicting wrong explicit
    answer, an \emph{answer-text-prioritizing} model will follow the wrong
    explicit answer (tracking the stated answer text), while a
    \emph{computation-prioritizing} model will follow its own computation and
    produce the correct answer (ignoring the stated answer text).
\end{hypothesis}
This third prediction directly opposes the two accounts: it cannot be
explained by format sensitivity or corruption artifacts, since the steps
themselves are intact and correct. Section~\ref{sec:conflicting} tests
this prediction using a three-condition causal experiment.

\section{Experimental Setup}
\label{sec:setup}

\subsection{Task Slices}
\label{sec:slices}

We evaluate four task slices that vary in chain format, difficulty, and
scale (Table~\ref{tab:slices}), plus a cross-domain commonsense set.

\paragraph{Primary versus exploratory analysis.}
The three core pre-specified tests are H1 (answer-line ablation collapses suffix
sensitivity), H2 (explicit answer text dominates behavioral output via conflicting-answer
protocol), and H3 (positional sensitivity tracks answer placement bidirectionally).
Results on additional models, datasets, and subgroups are clearly labeled as
replication, sensitivity, or exploratory in the corresponding sections.
All primary $p$-values are pre-specified one-per-endpoint; where multiple
comparisons appear within a section, we report Bonferroni-aware interpretations
alongside nominal values and label $0.05 < p < 0.10$ as directional.

\begin{table}[t]
  \centering
  \small
  \caption{Task slices used in this study. ``Answer in suffix'' indicates
  whether the chain's final steps explicitly restate the answer.}
  \label{tab:slices}
  \begin{tabular}{lccl}
    \toprule
    Slice & $N$ & Answer in suffix? & Purpose \\
    \midrule
    Hard-v3 (synthetic) & 60 & No & Prefix-sensitivity test \\
    GSM8K-v1 (benchmark) & 100 & Yes & Suffix-sensitivity test \\
    GSM8K-stripped-v1 & 300 & Removed & Format ablation \\
    GSM8K-conflict-v1 & 500 & Conflicting & Direct causal test \\
    Commonsense-v1 & 150 & Conflicting & Cross-domain generalization \\
    MATH-v1 (benchmark) & 100 & Yes (\textbackslash boxed) & Cross-domain format ablation \\
    MATH-stripped-v1 & 100 & Removed & MATH format ablation \\
    \bottomrule
  \end{tabular}
\end{table}

\paragraph{Hard-v3 (synthetic, $N{=}60$).}
A generated 60-example slice designed to suppress question-only shortcuts.\footnote{$N{=}60$ is the question-only-controlled subset: examples where the
question-only accuracy across three probing runs is $\leq$50\%. See
\S\ref{sec:controls} for the full QO-control procedure.}
Early chain steps explicitly extract relevant quantities into named symbolic
variables; middle steps operate on those variables; suffix steps do \emph{not}
directly restate the final answer. The answer must be derived from symbolic
anchors established in the prefix. The slice spans multiple problem domains
with varied distractors. Corrupted steps were generated by GPT-4o with
prompt-injection of wrong intermediate values (arithmetic errors), then
manually verified to ensure grammatical coherence and absence of correct-answer
leakage. The slice was frozen before model evaluation and has not been modified
post-hoc. Hard-v3 is used as \emph{corroborating evidence} for the
format-determination claim; the primary evidence is the within-GSM8K format
ablation (\S\ref{sec:ablation}) on naturally occurring benchmark chains.

\paragraph{GSM8K-v1 (benchmark, $N{=}100$).}
A deterministically filtered 100-example subset of the GSM8K test
set~\citep{cobbe2021}. We retain examples with integer-normalized final answers
and parsed gold rationales of at least four steps. Gold rationales
are segmented into prefix, middle, and suffix thirds. Critically, GSM8K gold
chains typically \emph{end} with an explicit answer statement (``The answer is
$X$''), making the suffix structurally distinct from the synthetic slice.

\paragraph{GSM8K-stripped-v1 (format ablation, $N{=}300$).}
GSM8K-v1 examples ($N{=}300$) with explicit answer statements
(e.g., ``The answer is 24'') removed from chain suffixes. This creates a
matched pair: identical questions, identical intermediate reasoning, but
the suffix no longer carries the answer verbatim. If positional sensitivity
shifts away from the suffix on this slice, the format-determination
hypothesis (Hypothesis~\ref{hyp:format}) is confirmed.
For Qwen~2.5-3B, we use $N{=}300$ for the matched format-ablation comparison
(Table~\ref{tab:ablation}) and $N{=}100$ for the cross-format comparison
(Table~\ref{tab:main-results}).

\paragraph{Commonsense-v1 (cross-domain, $N{=}150$).}
A set of 150 commonsense reasoning examples spanning five domains
(social, temporal, counterfactual, spatial, and causal reasoning).
Unlike the arithmetic slices, answers are text-based (e.g., ``grateful,''
``the fire grows''), testing whether the conflicting-answer effect
generalizes beyond numeric extraction.

\subsection{Models}
\label{sec:models}

We evaluate ten open-weight instruction-tuned models spanning five model families/lines
(Qwen~2.5, Qwen3, Phi, Mistral, and a DeepSeek-distilled Qwen-architecture variant):
\begin{itemize}
  \item \emph{Qwen\;2.5-3B-Instruct} (\texttt{Qwen/Qwen2.5-3B-Instruct}), 3.1B parameters
  \item Phi-3-mini-4k-instruct
    (\texttt{microsoft/Phi-3-mini-4k-instruct}), 3.8B parameters
  \item \emph{Qwen\;2.5-7B-Instruct} (\texttt{Qwen/Qwen2.5-7B-Instruct}), 7.6B parameters (GSM8K-v1 scale evaluation; Appendix~\ref{sec:samplesize_app})
  \item \emph{DeepSeek-R1-Distill-Qwen-7B} (\texttt{deepseek-ai/DeepSeek-R1-Distill-Qwen-7B}), 7B parameters; Qwen-2.5-Math-7B fine-tuned on DeepSeek-R1's reasoning traces (distillation hypothesis probe; Section~\ref{sec:14b})
  \item \emph{Mistral-7B-Instruct-v0.3} (\texttt{mistralai/Mistral-7B-Instruct-v0.3}), 7.2B parameters; third model family (Appendix~\ref{app:extended})
  \item \emph{Qwen\;2.5-14B-Instruct} (\texttt{Qwen/Qwen2.5-14B-Instruct}), 14.7B parameters (scale evaluation; Section~\ref{sec:14b})
  \item \emph{Phi-4} (\texttt{microsoft/phi-4}), 14B parameters; cross-family 14B scale replication (Section~\ref{sec:14b})
  \item \emph{Qwen\;2.5-32B-Instruct} (\texttt{Qwen/Qwen2.5-32B-Instruct}), 32.8B parameters (32B scale point; Section~\ref{sec:14b})
  \item \emph{Qwen3-8B} (\texttt{Qwen/Qwen3-8B}), 8B parameters; next-generation Qwen model.
  \item \emph{Qwen3-14B} (\texttt{Qwen/Qwen3-14B}), 14B parameters; within-Qwen3-generation scale gradient (Section~\ref{sec:14b}).
\end{itemize}
Phi-3-mini, Mistral, and Phi-4 provide cross-family replication: the override
effect holds across Qwen, Phi, and Mistral architectures, ruling out
single-family artifacts. DeepSeek-R1-Distill-7B extends the probe to the
training-objective axis: it is a Qwen-architecture model fine-tuned on
DeepSeek-R1's 671B reasoning traces, testing whether reasoning-oriented
distillation might transfer resistance to answer-text override (we do not observe this in the current single-pair comparison).
The 7B, 14B, and 32B evaluations establish a monotonically decreasing
scale-dependent gradient: CC accuracy drops to zero or near-zero ($\leq$~0.02)
at 7B across three model families (Qwen~2.5, Mistral, DeepSeek-distilled)
(FW strict$\,{=}\,0.30$; corrected$\,{=}\,$\SBCCFW),
dramatically attenuated at 14B (Phi-4: FW\,$=$\,\PHIFOURCCFW), and
near-zero at 32B (Qwen-32B: FW\,$=$\,\QWTBCCFW).

\subsection{Protocol}
\label{sec:protocol}

Each model is evaluated in five conditions per slice: chain-enabled baseline,
prefix-corrupted, middle-corrupted, suffix-corrupted, and question-only.
Semantic corruption replaces steps with syntactically valid but arithmetically
or logically incorrect alternatives: arithmetic operators are swapped
(e.g.\ $+\!\to\!-$, $\times\!\to\!\div$), numbers are incremented by one,
and common math phrases are exchanged (``plus''$\to$``minus'').
The corrupted step retains its original length and syntactic
structure, changing only the mathematical content.
Corruption fraction is 100\% of eligible
steps within the target region to maximize causal signal at these sample
sizes. Answer extraction uses a structured step that prefers a clean
leading integer line and evaluates simple arithmetic expressions when needed.

Statistical tests use exact paired sign tests per-example and bootstrap 95\%
confidence intervals (2000 resamples) for accuracy differences. A
pre-specified inferential hierarchy fixes the primary endpoints as
(H1)~conflicting-chain CC accuracy and (H2)~follow-wrong rate (FW), with
positional deltas (H3) treated as diagnostic; full hierarchy, extraction
policy, and multiplicity corrections are detailed in
Appendix~\ref{appendix:extraction}.
All primary evaluations use greedy decoding (temperature $= 0$, no sampling)
for reproducibility; temperature and prompt-wording sensitivity remain
open directions discussed in \S\ref{sec:limitations}.

\begin{figure}[!htbp]
  \centering
  \includegraphics[width=\linewidth]{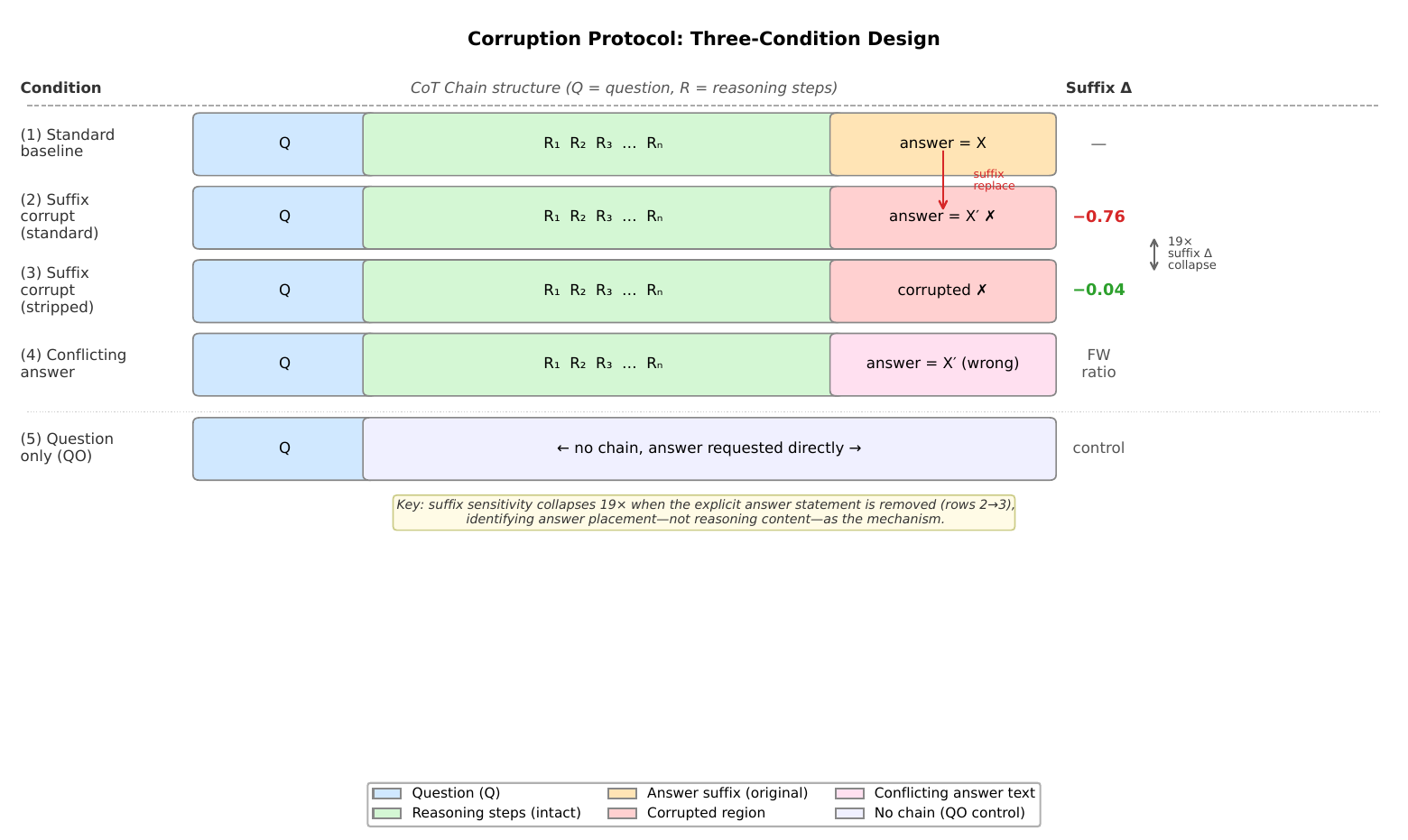}
  \caption{Protocol schematic: five-condition experimental design. Each question is
  evaluated in five conditions: (1)~standard-chain baseline, (2)~suffix-corrupted
  standard chain, (3)~suffix-corrupted stripped chain (answer statement removed),
  (4)~conflicting-answer chain, and (5)~question-only (no chain). The 19$\times$
  collapse of suffix sensitivity between rows~2 and~3 identifies answer placement
  as the mechanism driving positional sensitivity. The conflicting-answer condition
  (row~4) measures whether the model follows the wrong explicit answer or recovers
  the correct computation; FW$-$QO isolates chain-attribution beyond a question-only
  baseline (row~5).}
  \label{fig:protocol_schematic}
\end{figure}

\section{Core Experiment 1: Format Ablation}
\label{sec:ablation}

The format ablation (GSM8K-stripped-v1) tests Hypothesis~\ref{hyp:format}
by direct intervention on the chain format. Starting from the GSM8K-v1
chains, we programmatically remove the final sentence of each chain's
suffix when it matches the pattern ``The answer is \textit{[number]}''.
The remaining chain steps are left intact. If suffix sensitivity
disappears or diminishes on the stripped version while remaining strong
on the original, and if prefix or middle sensitivity increases, this
constitutes direct evidence that the positional signal was driven by
answer placement rather than computational structure.

This is a within-subject design: each example serves as its own control
across the two formats, eliminating confounds from question difficulty
or model capability.

\paragraph{GSM8K-conflict-v1 (causal test, $N{=}500$).}
A 500-example dataset derived from GSM8K-v1 and designed to directly
test Hypothesis~\ref{hyp:conflict}. Each example contains a chain whose
intermediate reasoning steps are correct (identical to \textsc{sc} steps),
but whose final explicit answer statement is replaced with a wrong
answer $a^-_x \neq a^*_x$. The wrong answer is chosen to be a plausible
wrong value (e.g., the true answer manipulated by a small arithmetic
perturbation). This creates a direct conflict between what the steps compute
and what the final answer text states. A more detailed description of the
three evaluation conditions is given in Section~\ref{sec:conflicting}.
We additionally evaluate a 200-example GSM8K-v2 subset in a
computation-terminal stress test: the final answer line is removed from the
correct chain to verify the model can still compute from intermediate steps,
then replaced with a stronger conflicting suffix to test whether explicit
answer text overrides that computation.

\section{Core Experiment 2: Reasoning $\times$ Answer-Line Factorial}
\label{sec:factorial}

\noindent
This experiment provides orthogonal evidence for answer-text dominance:
if models are sensitive to reasoning quality, Condition~B (correct reasoning $+$
wrong answer) should outperform Condition~C (corrupted reasoning $+$ correct
answer). Answer-text dominance predicts the opposite.

\subsection{Design and Predictions}

We evaluate four chain conditions, each on $N{=}\FSN$ GSM8K examples ($N_{\text{total}}{=}1{,}200$ across all four conditions):
\begin{itemize}
  \item \textbf{A} (correct reasoning + correct answer): standard chain
    accuracy serves as the ceiling.
  \item \textbf{B} (correct reasoning + wrong answer): our existing
    conflicting-answer condition; accuracy measures how often the
    model overrides its reasoning.
  \item \textbf{C} (wrong reasoning + correct answer): corrupted
    intermediate steps, but correct answer text in the final line.
    If the answer line rescues accuracy despite bad reasoning, this is
    direct evidence that answer text, not intermediate computation, drives
    the final output.
  \item \textbf{D} (wrong reasoning + wrong answer): corrupted steps +
    wrong answer; lowest-performing condition expected.
\end{itemize}

\paragraph{Decision predictions.}
Under \emph{reasoning dominance}: $A \approx B > C \approx D$; reasoning quality
determines accuracy and the answer line doesn't override it.

Under \emph{answer-text dominance}: $A \approx C > D \approx B$; the answer line
determines accuracy regardless of reasoning quality. Correct answer text
in~C rescues the model to near-standard performance despite corrupted steps.

Under a \emph{mixed} account: $A > C > D > B$  (or similar partial ordering)
with intermediate effects for both factors.

\subsection{Results}

\begin{table}[h]
  \centering
  \small
  \caption{2$\times$2 factorial: model accuracy (fraction producing correct answer)
  vs.\ followed-wrong rate for wrong-answer conditions.
  Both scales GSM8K ($N{=}\FSN$ per condition at 3B; $N{=}\FSeightN$ per condition at 8B).}
  \label{tab:factorial}
  \begin{tabular}{@{}lllcc@{}}
    \toprule
    Model & & & Correct answer-line & Wrong answer-line \\
          & & & \small(accuracy) & \small(accuracy | fw-rate) \\
    \midrule
    \multicolumn{5}{@{}l}{\small\textit{Qwen~2.5-3B-Instruct}} \\
    \addlinespace[2pt]
    & \multirow{2}{*}{Reasoning} & Correct & A: \FSCA & B: \FSCBACC{} | fw=\FSCB \\
    &                             & Wrong   & C: \FSCC & D: \FSCDACC{} | fw=\FSCD \\
    \addlinespace[4pt]
    \multicolumn{5}{@{}l}{\small\textit{Qwen3-8B-Instruct}} \\
    \addlinespace[2pt]
    & \multirow{2}{*}{Reasoning} & Correct & A: \FSeightCA & B: \FSeightCBacc{} | fw=\FSeightCB \\
    &                             & Wrong   & C: \FSeightCC & D: \FSeightCDacc{} | fw=\FSeightCD \\
    \bottomrule
  \end{tabular}
\end{table}

Corrupted-reasoning chains with a correct explicit answer
line (Condition~C) achieve accuracy $\FSCC$ at 3B and $\FSeightCC$ at 8B, within
a few percentage points of gold-reasoning chains with the same correct answer
line (Condition~A: $\FSCA$/$\FSeightCA$), showing that the correct answer statement
largely rescues accuracy despite corrupted intermediate steps.
Conversely, correct-reasoning chains with a wrong explicit answer
line (Condition~B) achieve only $\FSCBACC$ accuracy (fw=$\FSCB$) at 3B and
$\FSeightCBacc$ accuracy (fw=$\FSeightCB$) at 8B, well below their reasoning
potential. The dominant pattern aligns with \emph{answer-text primary} at both
scales: whether reasoning is corrupted matters far less ($\Delta\text{acc}
\approx 0.01$--$0.05$) than whether the answer line is correct
($\Delta\text{acc} \approx 0.56$--$0.94$).

\paragraph{Cross-scale convergence.} The full 2$\times$2 factorial now replicates
across Qwen~2.5-3B ($N{=}\FSN$/cell) and Qwen3-8B ($N{=}\FSeightN$/cell).
The answer-text primary ordering $A{\approx}C \gg B{>}D$ holds at both scales,
and the answer-line effect dwarfs the reasoning effect at 8B
(answer-line $\Delta$acc $\approx$0.94 vs.\ reasoning $\Delta$acc $\approx$0.01).
The B-cell followed-wrong rate increases from \FSCB{} at 3B to \FSeightCB{} at 8B,
and the C-cell rescue rate improves from \FSCC{} to \FSeightCC,
indicating that answer-text dominance is not attenuated at larger scale.
Critically, scaling from 3B to 8B does not recover reasoning-governed behavior:
correct reasoning reduces wrong-answer following by only $\sim$27~pp at 8B
(Condition~B vs.~D: fw=\FSeightCB{} vs.~\FSeightCD), confirming partial
protection rather than full reasoning override at this scale range.


\subsection{Question-Only Controls Detect Systematic Confounds}
\label{sec:controls}

Preliminary experiments on an easy arithmetic-comparison slice showed that
instruction-tuned models answer a large fraction of examples correctly from the
question alone. In this regime, robustness to corruption is uninformative: the
model bypasses the chain entirely.

On all hard task slices, we first verify the control requirement before
interpreting positional effects. Table~\ref{tab:main-results} reports the
complete results.

\begin{table}[!t]
  \centering
  \small
  \caption{Corruption results across slices and models. ``QO'' = question-only
  (no chain). Bold = largest accuracy drop from baseline.
  The positional sensitivity reverses completely with chain format:
  Hard-v3 (no answer-bearing suffix) is prefix-sensitive;
  GSM8K-v1 (explicit answer in suffix) is suffix-sensitive.
  The within-dataset format ablation (Table~\ref{tab:ablation}) isolates the
  mechanism; this table shows the cross-format reversal.}
  \label{tab:main-results}
  \begin{tabular}{llccccc}
    \toprule
    Slice & Model & Baseline & Middle & Prefix & Suffix & QO \\
    \midrule
    \multirow{2}{*}{Hard-v3 ($N{=}60$)}
      & Qwen~2.5-3B  & 0.183 & 0.050 & \textbf{0.017} & 0.267 & 0.083 \\
      & Phi-3-mini    & 0.800 & 0.717 & \textbf{0.033} & 0.700 & 0.233 \\
    \midrule
    \multirow{2}{*}{GSM8K-v1 ($N{=}100$)}
      & Qwen~2.5-3B  & 0.970 & 0.970 & 0.980 & \textbf{0.210} & 0.060 \\
      & Phi-3-mini    & 0.340 & 0.450 & 0.320 & \textbf{0.130} & 0.160 \\
    \bottomrule
  \end{tabular}
\end{table}

\paragraph{Hard-v3 chain gaps.}
For Phi-3-mini, the chain gap is $\Delta_\mathrm{QO} = -0.567$
(95\% CI $[-0.717, -0.417]$, $p = 1.9 \times 10^{-8}$),
confirming that the chain is load-bearing for the stronger model.
For Qwen~2.5-3B, the chain gap is $\Delta_\mathrm{QO} = -0.100$
(95\% CI $[-0.217, +0.017]$, $p = 0.180$),
reflecting the model's low baseline accuracy (0.183) on these hard items.
The gap does not reach conventional significance at $N{=}60$, so this
entry is marked with $^\dagger$ in Table~\ref{tab:crossmodel} as
borderline for Definition~\ref{def:2}. The directional effect is
present (QO $<$ Base), and positional drops remain significant
($p{<}0.01$), but the QO prerequisite is formally borderline.

\paragraph{GSM8K-v1 chain gap.}
Question-only accuracy is 0.060, far below the 0.970 chain-enabled baseline:
$\Delta_\mathrm{QO} = -0.910$
(95\% CI $[-0.960, -0.850]$, $p = 1.9 \times 10^{-12}$).

\subsection{Position Sensitivity Is Format-Determined}
\label{sec:format-determined}

The central finding emerges from comparing positional effects across the
two chain formats (Table~\ref{tab:main-results}).
We note that this cross-format comparison contrasts different task
slices (Hard-v3 vs.\ GSM8K-v1), so task difficulty and chain structure
are not equated. The format-determination mechanism is directly
established by the within-dataset format ablation (\S\ref{sec:ablation}),
which varies only the chain format while holding task, model, and examples
constant; Table~\ref{tab:main-results} shows the qualitative reversal that
motivates the ablation.

\paragraph{Hard-v3: prefix is load-bearing.}
On the synthetic slice where suffixes do not restate the answer:
\begin{itemize}
  \item Qwen~2.5-3B$^\dagger$ (descriptive only; QO borderline $p=0.180$):
  prefix $\Delta = -0.167$ ($p = 0.006$),
  middle $\Delta = -0.133$ ($p = 0.008$),
  suffix $\Delta = +0.083$ ($p = 0.125$).

  \item Phi-3-mini: prefix collapses by $\Delta_\mathrm{prefix} = -0.767$
  (95\% CI $[-0.867, -0.650]$, $p = 1.3 \times 10^{-12}$).
  Middle does not separate from baseline:
  $\Delta_\mathrm{middle} = -0.083$
  (95\% CI $[-0.200, +0.033]$, $p = 0.267$).
  Suffix shows a non-significant decrease ($\Delta_\mathrm{suffix} = -0.100$,
  95\% CI $[-0.217, +0.000]$, $p = 0.146$).
\end{itemize}

\paragraph{GSM8K-v1: suffix is load-bearing.}
On the benchmark slice where suffixes contain ``the answer is $X$'':
\begin{itemize}
  \item Qwen~2.5-3B: suffix corruption collapses accuracy to 0.210:
  $\Delta_\mathrm{suffix} = -0.760$
  (95\% CI $[-0.840, -0.670]$, $p = 1.4 \times 10^{-12}$).
  Middle corruption shows zero damage ($\Delta_\mathrm{middle} = 0.000$, $p = 1.0$).
  Prefix corruption likewise shows no damage ($\Delta_\mathrm{prefix} = +0.010$, $p = 1.0$).

  \item Phi-3-mini: suffix corruption collapses accuracy from 0.340 to 0.130:
  $\Delta_\mathrm{suffix} = -0.210$
  (95\% CI $[-0.300, -0.120]$, $p = 1.9 \times 10^{-5}$).
  Prefix corruption shows no effect ($\Delta_\mathrm{prefix} = -0.020$, $p = 0.83$).
  Middle corruption shows a non-significant increase
  ($\Delta_\mathrm{middle} = +0.110$, $p = 0.08$).
\end{itemize}

\paragraph{The reversal.}
The same corruption protocol, the same statistical tests, yet
the ``causally critical'' position flips from prefix to suffix. The only
difference is the chain format. On hard-v3, where the prefix establishes
symbolic anchors and the suffix does not restate the answer, prefix
corruption is catastrophic. On GSM8K-v1, where the suffix carries the
explicit answer statement, suffix corruption is catastrophic instead.
The primary cross-model evidence comes from Phi-3-mini, which satisfies
the QO prerequisite with high confidence on both slices; Qwen~2.5-3B
shows the same directional pattern on Hard-v3 (descriptive only, given
the borderline QO gap) and the same pattern on GSM8K-v1 where the
prerequisite is fully satisfied ($p < 10^{-12}$).
This reversal directly contradicts any universal claim about
which chain positions are computationally important and motivates
Hypothesis~\ref{hyp:format}: corruption studies detect answer placement,
not computational depth.

\paragraph{Primary vs.\ secondary/converging evidence.}
For readability, the evidence hierarchy is:
\begin{itemize}[nosep]
  \item \emph{Primary} (protocol-uniform, no post-hoc subset conditioning):
    the 3B within-dataset format ablation
    ($N{=}300$, $19\times$ collapse) and the conflicting-answer CC-accuracy
    collapse at 7B ($\leq 0.02$ across three model families,
    extraction-invariant).
  \item \emph{Secondary / converging}: the 7B within-stable subset
    ($N{=}76$, $9.3\times$ attenuation; post-hoc conditioning), the
    Qwen3-8B within-stable replication ($N{=}299$, $p{=}0.004$), the
    MATH replication, answer-placement controls, commonsense, self-generated
    chains, base-model comparison, and cross-branch scale syntheses.
\end{itemize}
Secondary/converging results are robustness checks that point in the same
direction as the primary endpoints; they should not be read as independent
confirmatory tests.

\begin{table}[t]
  \centering
  \small
  \caption{Primary confirmatory endpoints. Protocol-uniform designs with
    matched examples, identical extraction, and no post-hoc subset conditioning.
    $\dagger$~Format ablation: same examples, same reasoning, only answer
    statement removed. $\ddagger$~Conflicting-answer: correct reasoning with wrong
    terminal answer; extraction-invariant. All $p$-values from paired exact tests
    (McNemar for ablation, binomial for conflicting-answer).}
  \label{tab:primary_endpoints}
  \begin{tabular}{llcccccl}
    \toprule
    Test & Model & $N$ & SC Acc & CC Acc / $\Delta_{\text{suf}}$ & FW Rate & QO & $p$ \\
    \midrule
    \multicolumn{8}{@{}l}{\textbf{Format Ablation (answer statement removed)}$^{\dagger}$} \\
    \cmidrule(l){1-8}
    & Qwen~2.5-3B$^{\ddag}$  & 300 & 1.000 & $\Delta{=}{-}0.040$ & --- & 0.060 & 0.022 \\
    & Qwen~2.5-7B$^{\ddag}$  & 300 & 0.273 & $\Delta{=}{+}0.117$ & --- & 0.10$^{*}$ & ${<}10^{-5}$ \\
    & Qwen3-8B$^{\ddag}$     & 299 & 0.997 & $\Delta{=}{-}0.027$ & --- & 0.010 & 0.004 \\
    & Phi-3-mini$^{\ddag}$   & 100 & 0.900 & $\Delta{=}{+}0.020$ & --- & 0.160$^{*}$ & $1.0^{\natural}$ \\
    & DeepSeek-R1-7B (MATH)   & 100 & 0.540 & $\Delta{=}{-}0.030$ & --- & 0.010$^{*}$ & $1.0^{\natural}$ \\
    \midrule
    \multicolumn{8}{@{}l}{\textbf{Conflicting-Answer (answer-text override)}$^{\ddagger}$} \\
    \cmidrule(l){1-8}
    & Qwen~2.5-3B$^{\S}$  & 500 & 0.970 & 0.280 & 0.630 & 0.060 & ${<}10^{-8}$ \\
    & Qwen~2.5-7B$^{\S}$  & \SBN{} & 0.990 & 0.000 & 1.000 & 0.050 & ${<}10^{-30}$ \\
    & Mistral-7B$^{\S}$   & 500 & 1.000 & 0.020 & 0.980 & 0.050 & ${<}10^{-30}$ \\
    & Phi-4-14B$^{\P}$    & 100 & 0.850 & 0.390 & 0.300 & 0.000 & ${<}10^{-8}$ \\
    & Qwen~2.5-32B$^{\P}$ & 100 & 0.960 & 0.940 & 0.010 & 0.300 & $1.0^{\natural}$ \\
    \bottomrule
  \end{tabular}
  \vspace{4pt}
  {\footnotesize
  $^{\ddag}$~Neutral-stripped format (answer replaced with neutral placeholder).\\
  $^{\S}$~GSM8K-v2 strong-suffix format; extraction-invariant CC accuracy.
  Qwen~2.5-3B and Mistral-7B: $N{=}500$; Qwen~2.5-7B: $N{=}\SBN$ (same rows as Table~\ref{tab:conflicting}).\\
  $^{\P}$~GSM8K-v1 format ($N{=}100$).\\
  $^{*}$~QO measured on same dataset.\\
  $^{\natural}$~Non-significant result (expected under format-determination at these scale points).\\
  }
\end{table}

\bigskip
\noindent\textbf{Core Evidence Roadmap.}
Table~\ref{tab:primary_endpoints} lists the pre-specified primary endpoints.
The paper now presents four core experiments of increasing causal strength,
each testing a progressively sharper prediction:
\begin{enumerate}[nosep]
  \item \emph{Core Experiment 1: Format Ablation (\S\ref{sec:ablation}):} Same model, same
    examples, same reasoning, only the answer statement removed. Tests whether
    positional sensitivity tracks answer placement.
  \item \emph{Core Experiment 2: Reasoning $\times$ Answer-Line Factorial (\S\ref{sec:factorial}):}
    Crosses reasoning correctness with answer-line content. Tests whether answer
    text dominates reasoning quality in a $2\times2$ design.
  \item \emph{Core Experiment 3: Conflicting-Answer Direct Override (\S\ref{sec:conflicting}):}
    Correct reasoning with wrong terminal answer. Tests whether models follow
    answer text over their own computation.
  \item \emph{Core Experiment 4: Answer-Placement Controls (\S\ref{sec:placement_controls}):}
    Answer relocation, bidirectional ablation, and counterbalanced positioning.
    Tests whether the critical variable is answer content at the readout position.
\end{enumerate}
\bigskip

\subsection{Format Ablation: Results and Mechanism}
\label{sec:ablation_results}

\begin{table*}[t]
  \centering
  \small
  \caption{Format ablation. GSM8K-v1 (original, with ``the answer is $X$'') vs.\ GSM8K-neutral-stripped-v1
  (answer statement replaced with neutral placeholder). Within-stable comparison
  for Qwen~2.5-7B: $9.3\times$ suffix-sensitivity attenuation ($N{=}76$, $p{=}7.8{\times}10^{-3}$).
  Full $N{=}300$ inverts from $-0.643$ to $+0.117$ but is directional under baseline-drop caveat.
  See footnotes for protocol details.}
  \label{tab:ablation}
  \begin{tabular}{llccccc}
    \toprule
    Format & Model & Baseline & Middle & Prefix & Suffix & QO \\
    \midrule
    \multirow{4}{*}{GSM8K-v1 (original)}
      & Qwen~2.5-3B  & 0.970 & 0.970 & 0.980 & \textbf{0.210} & 0.060 \\
      & Phi-3-mini    & 0.340 & 0.450 & 0.320 & \textbf{0.130} & 0.160 \\
      & Qwen~2.5-7B$^\ddagger$   & 0.606 & 0.523 & 0.612 & \textbf{0.011} & 0.10$^{**}$   \\
      & Qwen~2.5-14B  & 0.930 & 0.930 & 0.930 & \textbf{0.760} & ---$^\|$ \\
    \midrule
    \multirow{2}{*}{GSM8K-stripped-v1 (verify)}
      & Qwen~2.5-3B$^\dagger$  & 0.960 & \textbf{0.897} & 0.943 & 0.920 & 0.060$^*$ \\
      & Phi-3-mini    & 0.410 & 0.450 & \textbf{0.430} & 0.530 & 0.160$^*$ \\
    \midrule
    \multirow{3}{*}{GSM8K-neutral-stripped-v1}
      & Qwen~2.5-7B$^\diamond$ & 0.273 & \textbf{0.243} & 0.310 & 0.390$^\uparrow$ & 0.10$^{*{**}}$ \\
      & Phi-3-mini   & 0.900 & 0.910$^\uparrow$ & 0.920$^\uparrow$ & 0.920$^\uparrow$ & 0.160$^*$ \\
      & Qwen~2.5-14B & 0.900 & \textbf{0.870} & 0.900 & 0.880$^\S$ & ---$^\|$ \\
      & Qwen~2.5-32B & 1.000 & \textbf{0.980} & 1.000 & 1.000 & 0.300$^*$ \\
      & Qwen3-8B-Instruct ($N{=}300$)$^\P$ & \QTHREENSBASE & \QTHREENSMID & \QTHREENSPRE & \textbf{\QTHREENSSUF} & \QTHREENSQO \\
    \midrule
    \multirow{1}{*}{MATH-v1 (original)}
      & DeepSeek-R1-7B & \MATHORIGbase & \MATHORIGmiddle & \MATHORIGprefix & \textbf{\MATHORIGsuffix} & \MATHQO \\
    \midrule
    \multirow{1}{*}{MATH-stripped-v1}
      & DeepSeek-R1-7B & \MATHSTRIPbase & \textbf{\MATHSTRIPmiddle} & \MATHSTRIPprefix & \MATHSTRIPsuffix & \MATHQO$^*$ \\
  \end{tabular}

  \vspace{4pt}
  {\footnotesize
  $\uparrow$ Suffix corruption \emph{improves} accuracy (corruption removes suppressive placeholder signal).\hfill
  $^*$ GSM8K-stripped uses same questions as GSM8K-v1; QO identical across formats.\\
  $^{**}$ Qwen~2.5-7B QO directly measured ($N{=}100$; 10/100).\hfill
  $^\diamond$ Qwen~2.5-7B neutral-stripped: $N{=}300$, suffix $\Delta{=}{+}0.117$ ($p{<}10^{-5}$), directional under baseline-drop caveat.\\
  $^\ddagger$ Qwen~2.5-7B GSM8K-v1: $N{=}1{,}000$.\hfill
  $^\dagger$ Qwen~2.5-3B GSM8K-stripped-v1: $N{=}300$; all other rows $N{=}100$ unless noted.\\
  $^\|$ QO not measured for this row.\hfill
  $^\P$ Qwen3-8B-Instruct ($N{=}300$), thinking mode; suffix $\Delta{=}{-}0.027$ vs.\ prefix; within-stable $N{=}299$, McNemar $p{=}0.004$.\\
  $^\S$ $8.5\times$ smaller than standard-format suffix for same model ($\Delta{=}{-}0.17$, $p{=}0.001$).}
\end{table*}

The format ablation is the most direct test of Hypothesis~\ref{hyp:format}.
By removing only the explicit answer statement from GSM8K suffixes while
preserving all intermediate reasoning, we isolate the contribution of
answer placement to positional sensitivity.

Table~\ref{tab:ablation} presents the ablation results alongside the
original GSM8K-v1 numbers. The prediction under Hypothesis~\ref{hyp:format}
is clear: if suffix sensitivity on GSM8K-v1 is driven by the ``the answer
is $X$'' statement, then removing that statement should reduce suffix
sensitivity. If instead the suffix carries genuine computation that
happens to coincide with the answer statement, stripping the answer text
should have minimal effect on the positional pattern.

Results for Qwen~2.5-3B on GSM8K-stripped-v1 ($N{=}300$) confirm the
prediction: baseline accuracy is 0.960. Suffix corruption
produces only $\Delta_\mathrm{suffix} = -0.040$ ($p = 0.022$, two-sided sign test; 18 degradations, 6 improvements), a
dramatic 19$\times$ reduction from the $\Delta_\mathrm{suffix} = -0.760$
observed on the original GSM8K-v1 format, confirming
($p{=}0.022$, $N{=}300$) that suffix sensitivity on the original was
substantially driven by answer placement.
The Qwen~2.5-7B neutral-stripped results (reported directly below)
provide significant confirmation via within-stable $9.3\times$ attenuation
($N{=}76$, $p{=}7.8\times 10^{-3}$), with full-sample sign inversion as
directional corroboration;
taken together these constitute multi-scale converging evidence for
the format-determination hypothesis (see also \S\ref{sec:analysis}
for the formal claim-by-claim evaluation).

\emph{Ceiling and extraction.} Whether the baseline rise reflects improved
extraction or genuine accuracy, the key comparisons are unaffected:
$\Delta$ values compare corrupted-vs-baseline within the same format, so
a uniform extraction offset cancels identically. The 19$\times$ ratio is
a within-$\Delta$ comparison, not a raw-accuracy comparison.
After stripping, no single position dominates; the largest drop shifts to
middle corruption ($\Delta = -0.063$, $p < 0.001$; Bonferroni $p < 0.001$).

\paragraph{Phi-3-mini stripped: second-order format artifact.}
For Phi-3-mini, the stripped condition introduces a new artifact:
the synthetic placeholder (``Therefore.\ Let me verify this computation.'')
triggers sign-negation in 58/100 baseline predictions, yielding
$\Delta_\text{suffix}{=}+0.120$ ($p=0.012$) from disrupted sign-flip cues.
This is a placeholder-format artifact, not a replication of the
format-determination hypothesis; we report it as a caution about synthetic
format manipulations.

\paragraph{Qwen~2.5-7B stripped: verify-placeholder artifact at scale.}
The same placeholder causes 94/100 sign-negations at 7B (baseline $= 0.06$).
Corrupting the suffix improves accuracy to 0.16
($\Delta_\text{suffix}{=}+0.10$, $p{=}0.013$), confirming the placeholder
text, not generic failure, is the operative cause.

\paragraph{Qwen~2.5-7B neutral-stripped: within-stable attenuation at 7B.}
Replacing the verify-placeholder with ``The calculation above gives the
result.'' eliminates the extreme artifact (baseline: $0.060 \to 0.273$),
which entails a substantial baseline drop (55\% relative loss, $0.606 \to 0.273$; $N{=}300$).
Because this baseline collapse confounds full-sample effect estimates, the
primary 7B evidence is the within-stable-subset analysis ($N{=}76$),
which conditions on examples answered correctly with or without the answer line.

\emph{Within-stable-subset analysis (secondary/converging evidence).}
Restricting to the $N{=}76$ examples that are stably correct under \emph{both}
formats (baseline${=}1.000$ in both conditions; 76 of 300 matched examples),
the standard format yields
$\Delta_\mathrm{suffix}{=}{-}0.974$ (baseline: 1.000 $\to$ corrupted: 0.026;
sign test: 74 degradations, 0 improvements, $p = 1.1 \times 10^{-22}$), while the
neutral-stripped format yields only $\Delta_\mathrm{suffix}{=}{-}0.105$ (0 improvements,
8 degradations; $p{=}7.8\times 10^{-3}$), a 9.3$\times$ attenuation of suffix sensitivity
from removing the explicit answer line.
These 76 examples are solved correctly \emph{regardless} of format, so
the baseline drop plays no role: the entire 9.3$\times$ effect is attributable
to the presence vs.\ absence of the explicit answer statement.
Prefix corruption has negligible effect in either format
($|\Delta_\text{prefix}|{<}0.03$).

\emph{Full-sample sign-inversion (corroborating directional evidence).}
Across all $N{=}300$ matched examples, suffix corruption under the
neutral-stripped format no longer degrades accuracy, instead, it \emph{significantly improves}
accuracy ($\Delta_\text{suffix}{=}{+}0.117$, $p{<}10^{-5}$, sign test: 45
improvements, 10 degradations). This sign reversal, from catastrophically
negative ($\Delta{=}{-}0.643$ under the original format, $N{=}300$) to significantly
positive, is directionally consistent with format-determination. However,
because the neutral-stripped baseline falls from $0.606$ to $0.273$, the
full-sample inversion is partly confounded by answer-text-dependent examples
that fail entirely without the explicit answer cue. We therefore treat it as
corroborating directional evidence, not the primary 7B effect estimate.
The conflicting-answer result (\S\ref{sec:conflicting}), CC accuracy
dropping to zero or near-zero ($\leq$~0.02) at 7B across three model families (Qwen~2.5-7B, Mistral-7B, DeepSeek-R1-Distill-7B), is the primary 7B
format-determination evidence because it is extraction-invariant and
independent of subset conditioning. The within-stable analysis here
provides complementary converging evidence from the format-ablation
direction, with the two approaches together constituting multi-method
confirmation.

\textbf{Evidence tier.} The conflicting-answer CC-accuracy-to-near-zero result
(Table~\ref{tab:primary_endpoints}; extraction-invariant on the listed GSM8K
slices, with 7B rows at $N{=}\SBN$ for Qwen~2.5-7B and $N{=}500$ for Mistral-7B)
is the primary 7B format-determination endpoint. The within-stable analysis here
($N{=}76$) is secondary/converging evidence: it points in the same direction
from the format-ablation side but relies on post-hoc subset conditioning.

\emph{Baseline drop analysis.}
A matched-example decomposition on $N{=}300$ matched pairs shows that 197 of 300 examples are standard-correct; of these, 121 fail entirely without the explicit answer cue, while 76 succeed under both formats (the within-stable-subset used in the $9.3\times$ attenuation analysis above). This baseline drop is \emph{predicted} by the format-determination
hypothesis, if the ``the answer is $X$'' cue is causally necessary for some
examples, removing it should harm accuracy, but it also confounds the
full-sample inversion as an effect-size estimate.
An alternative concern is that the placeholder replacement itself changes
prompt pragmatics broadly, not merely removing the answer.
\paragraph{EXP-29: Answer-relocation position-content control.}
The context-header relocation control (\S\ref{sec:relocation}; $N{=}\PCTXN$,
same examples) addresses this directly: relocating the wrong answer rather
than removing it preserves baseline accuracy ($\approx\PCTXSTDACC$), confirming
that baseline drop is caused by the loss of the answer signal specifically,
not by the placeholder text. Together, the stripped ablation and the relocation
control provide converging evidence that the decisive causal variable is
answer-text \emph{content}, not position or format style.

\paragraph{Qwen3-8B-Instruct neutral-stripped: largest within-stable comparison.}
The Qwen3-8B-Instruct (thinking mode) provides the largest within-stable
comparison in the paper ($N{=}299$ of 300 examples answered correctly under
both conditions), directly addressing the sample-size limitation of the 7B
within-stable analysis ($N{=}76$).
On the neutral-stripped format, prefix corruption has zero effect
($\Delta_\text{prefix}{=}0.000$; $299/299$ correct), while suffix corruption
produces a small but statistically significant degradation:
$291/299$ correct ($\Delta_\text{suffix}{=}{-}0.027$,
95\% CI $[-0.052, -0.014]$; exact McNemar vs.\ prefix: $b{=}8$, $c{=}0$,
$p{=}0.0039$).
On the standard GSM8K-v1 format ($N{=}100$), the same model shows zero degradation
at \emph{all} three positions (prefix, middle, and suffix all 100\%), confirming
that thinking-mode robustness does not eliminate positional residuals under the
neutral-stripped format.
This $N{=}299$ within-stable comparison provides protocol-uniform, high-power
evidence that format-determination persists at 8B scale even in thinking mode,
strongly consistent with the scale gradient observed from 3B through 14B.
Taken together, the 7B within-stable analysis ($N{=}76$, $9.3\times$ attenuation) and
the Qwen3-8B within-stable analysis ($N{=}299$, McNemar $p{=}0.0039$) provide
two-scale, protocol-uniform converging evidence for format-determination at the 7B and 8B
range, with the 8B result offering the largest within-stable sample in the paper.

\paragraph{Phi-3-mini neutral-stripped: cross-family format-determination.}
The same neutral-placeholder format applied to Phi-3-mini yields near-zero
sensitivity at all three positions
($\Delta_\text{middle}{=}{+}0.01$, $\Delta_\text{prefix}{=}{+}0.02$,
$\Delta_\text{suffix}{=}{+}0.02$, $N{=}100$;
Table~\ref{tab:ablation}).
Baseline accuracy rises from $0.340$ (original GSM8K, where explicit answer
text sometimes overrides reasoning) to $0.900$ (neutral-stripped, where the
model relies only on intermediate computation).
Taken together, the original $\Delta_\text{suffix}{=}{-}0.210$ collapses to
$+0.020$ on the neutral-stripped format, an approximate $10\times$ reduction
in magnitude with sign inversion, confirming the same qualitative
format-determination pattern seen in Qwen~2.5-3B (${\approx}19\times$ collapse)
and Qwen~2.5-7B ($9.3\times$ within-stable attenuation, $N{=}76$,
$p{=}7.8\times10^{-3}$; full-sample sign inversion as directional corroboration).
This cross-family replication rules out the possibility that
suffix-answer-text dominance is an artefact of the Qwen model family.

\paragraph{Cross-domain replication: MATH.}
To test whether the format-determination mechanism extends beyond arithmetic
word problems, we evaluated DeepSeek-R1-Distill-Qwen-7B on $N{=}\MATHN$
MATH competition problems using the same position-corruption protocol.
The question-only control confirms the prerequisite: QO accuracy is \MATHQO,
far below the chain-enabled baseline (original: \MATHORIGbase; stripped:
\MATHSTRIPbase), confirming that the chain is necessary for this model on
these problems.
On the original MATH format, where chains end with an explicit
``$\backslash$boxed\{$X$\}'' answer, suffix corruption is catastrophic
($\Delta_\text{suffix} = -0.630$; survival rate \MATHORIGsuffixFW), while
prefix and middle corruption are benign ($\Delta_\text{prefix} = +0.030$,
$\Delta_\text{middle} = -0.030$). On the stripped variant (answer removed,
neutral filler inserted), the pattern inverts: suffix corruption becomes
negligible ($\Delta_\text{suffix} = -0.030$; survival rate \MATHSTRIPsuffixFW), a $10.9\times$ recovery.
Middle becomes the most sensitive position
($\Delta_\text{middle} = -0.150$). This cross-domain replication on a
qualitatively different benchmark (competition mathematics vs.\ grade-school
arithmetic) with a different model (DeepSeek-R1 distillation vs.\ Qwen/Phi)
confirms that the format-determination mechanism is not benchmark-specific
or model-family-specific.

\paragraph{32B neutral-stripped: format-determination vanishes at scale.}
Qwen~2.5-32B-Instruct on the same neutral-stripped format ($N{=}100$)
achieves baseline accuracy 1.000, the model answers every question correctly
without any explicit ``the answer is $X$'' suffix (contrast 7B baseline$\,=\,$0.273 ($N{=}300$)
on the same format). Position-control corruption yields
$\Delta_\text{middle}{=}{-}0.020$, $\Delta_\text{prefix}{=}0.000$,
$\Delta_\text{suffix}{=}0.000$ (Table~\ref{tab:ablation}).
The near-zero sensitivity at all positions on the stripped format, combined
with the perfect baseline, indicates that 32B extracts its answer from the
intermediate reasoning steps alone. Format-determination, which persists
through 14B ($8.5\times$ sensitivity ratio), has effectively vanished at 32B.
This completes the scale gradient for both phenomena: answer-text override
(FW: $\QWSBVONEFW \to \QBVONEFW \to \QWTBCCFW$) and format-determination
(sensitivity ratio: $19\times$ at 3B, residual at 8B (within-stable $N{=}299$, McNemar $p{=}0.004$), $8.5\times$ at 14B, ${\approx}1\times$ at 32B)
both decline monotonically with scale.

\begin{figure}[!htbp]
  \centering
  \includegraphics[width=0.80\linewidth]{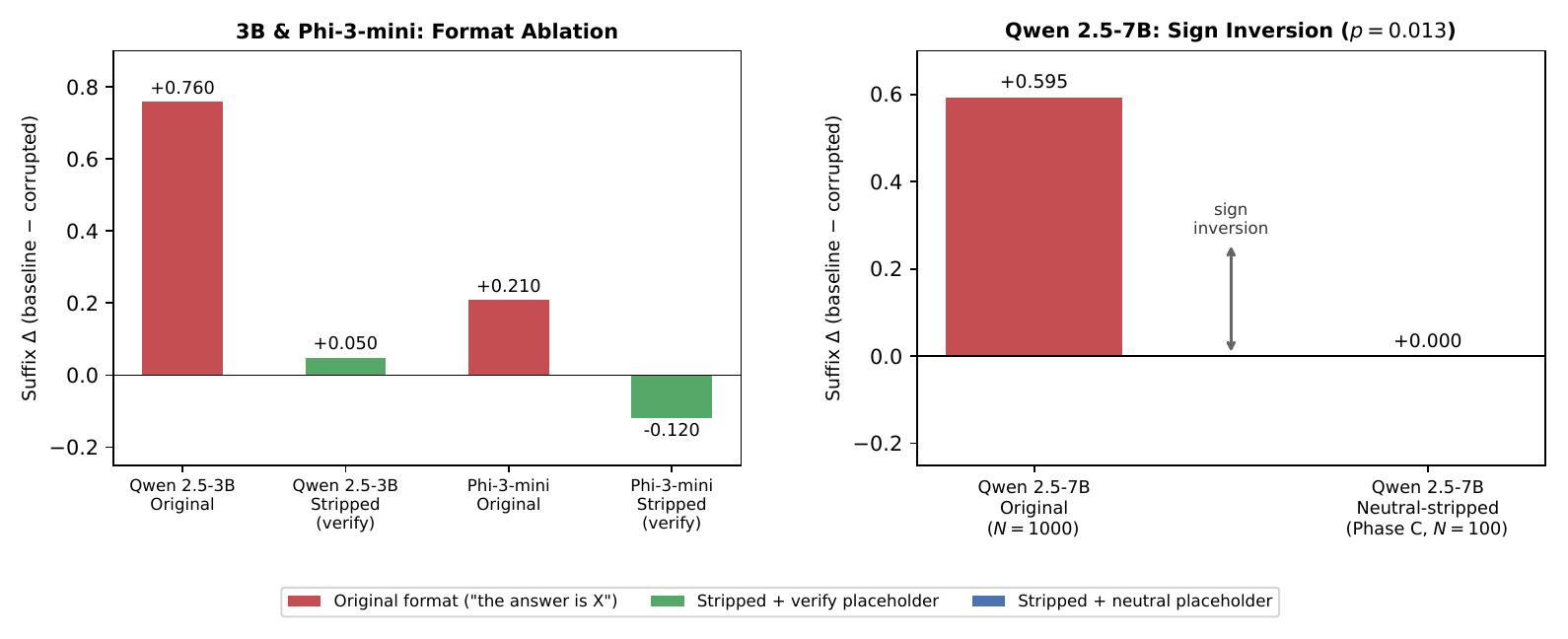}
  \caption{Format ablation: suffix sensitivity shrinks ${\approx}19\times$ when
  the explicit answer statement is removed from GSM8K suffixes
  (Qwen~2.5-3B: $\Delta_\text{suffix}$: $-0.760 \to -0.040$, $N{=}300$). This confirms the
  answer-placement mechanism for Qwen~2.5-3B. At the 7B scale, the clean
  within-stable comparison shows a $9.3\times$ attenuation ($N{=}76$,
  $p{=}7.8\times10^{-3}$); the full-sample neutral-placeholder point inverts
  from $-0.643$ (original GSM8K, $N{=}300$) to $+0.117$ ($p{<}10^{-5}$,
  $N{=}300$, neutral-stripped), but is directional because of the baseline drop.
  For Phi-3-mini, the same neutral-placeholder
  format confirms cross-family format-determination:
  $\Delta_\text{suffix}$ collapses from $-0.210$ (original) to $+0.020$ (neutral-stripped),
  a $10\times$ reduction with sign inversion ($N{=}100$ per condition;
  7B original $N{=}1{,}000$).}
  \label{fig:ablation}
\end{figure}

\section{Three-Prerequisite Protocol for Corruption Studies}
\label{sec:three_prereq}

The format-determination finding has a direct methodological consequence:
corruption-based causal analyses of CoT faithfulness require explicit
controls for answer-text placement before positional results can be
interpreted as evidence about computation. We propose three prerequisites
that should be standard for any such study.

\begin{enumerate}
  \item \emph{Question-only control.} Verify that chain-enabled accuracy
    significantly exceeds question-only accuracy ($p < 0.05$, paired test).
    If the gap is not significant, the model may bypass the chain entirely,
    and positional results are uninterpretable.
  \item \emph{Format characterization.} Determine where in the chain the
    final answer is explicitly encoded. Report this as metadata for the
    evaluation corpus. If the answer appears in a fixed position across
    examples (e.g., all chains end with ``the answer is $X$''), flag this as
    a potential confound.
  \item \emph{All-position sweep.} Test prefix, middle, and suffix
    corruption independently. A single-position study cannot distinguish
    answer-text sensitivity from computational dependence. Report results for
    all positions regardless of significance.
\end{enumerate}

\paragraph{Concrete application.}
For a benchmark where gold chains end with explicit answer statements (the
dominant format in GSM8K, MATH, and many instruction-tuned evaluation suites):
(a)~run the standard corruption sweep across all positions;
(b)~rerun the same sweep on a format-ablated version (answer statement
removed, neutral placeholder inserted);
(c)~compare the positional sensitivity patterns. If the dominant position
shifts when the answer statement is removed, the original finding was driven
by answer placement, not computation. Our experiments demonstrate this shift
at 3B (19$\times$ collapse), 7B ($9.3\times$ within-stable attenuation, with
full-sample inversion as directional corroboration), and across multiple model
families (Qwen, Qwen3, Phi; \S\ref{sec:ablation}).

\paragraph{Scope of applicability.}
The format-determination effect persists through 14B ($8.5\times$ sensitivity
ratio, $p{=}0.001$; Tab.~\ref{tab:ablation}; \S\ref{sec:14b}), even as
the direct override attenuates. At 32B, both effects converge toward zero
(Tab.~\ref{tab:ablation}: baseline$\,=\,$1.000, max $|\Delta|{=}0.020$
on neutral-stripped format). The protocol is therefore most critical at the
3B--14B scales where corruption conclusions are most vulnerable
to format artifacts, and remains useful as a diagnostic even when effects
are expected to be small.

\subsection{Extended Replications Summary}
\label{sec:crossmodel_summary}

The three-prerequisite protocol (\S\ref{sec:three_prereq}) applies at any
scale where the QO prerequisite is met. Across five model families
(Qwen~2.5, Qwen3, Phi, Mistral, and DeepSeek-distilled) and scales 3B--32B, the
conflicting-answer override is consistently present and extractable
(Table~\ref{tab:primary_endpoints}). Key findings from the full
cross-model and cross-scale analysis (Appendix~\ref{app:extended}):

\begin{itemize}[nosep]
  \item \emph{7B override is total across three model families.}
    CC accuracy $\leq$ 0.02 for Qwen-7B (0.00), DeepSeek-R1-Distill-7B
    (0.02), and Mistral-7B (0.02); extraction-invariant (magnitude-corrected
    and strict extraction both yield CC $\leq$ 0.02).
  \item \emph{Override attenuates monotonically with scale.}
    Followed-wrong rate drops from $\approx$1.00 at 7B to
    \PHIFOURCCFW\ at Phi-4-14B (\PHIFOURCCPVAL) to
    \QWTBCCFW\ at 32B.
  \item \emph{Format-determination persists as override fades.}
    At 14B, the $8.5\times$ sensitivity ratio between standard and
    neutral-stripped suffixes ($p{=}0.001$) confirms format-determination
    outlasts direct override.
  \item \emph{Both effects converge toward zero at 32B.}
    Neutral-stripped baseline reaches 1.000; max $|\Delta|{=}0.020$.
\end{itemize}

Full details, per-model statistics, extraction-policy comparisons, and
discussion of sign-negation artifacts are in Appendix~\ref{app:extended}.

\paragraph{Sample size.}
All primary endpoints use $N \geq 100$; GSM8K conflicting-answer runs use
$N{=}100$--$500$ depending on model and protocol branch (Table~\ref{tab:conflicting}):
the primary Qwen~2.5-3B and Mistral-7B GSM8K-v2 strong-suffix rows each use $N{=}500$,
whereas Qwen~2.5-7B uses $N{=}\SBN$ under the same
v2 construction; DeepSeek-R1-Distill-7B is reported at $N{=}\DSN$ on GSM8K-v1.
Paired exact tests reject at ${<}10^{-8}$ for all primary
comparisons. A 7B scale replication at $N{=}1{,}000$ confirms the positional
pattern is insensitive to sample size (Appendix~\ref{app:extended}).
\subsection{Core Experiment 3: Conflicting-Answer Direct Override}
\label{sec:conflicting}

\noindent
The cleanest causal test: present chains with \emph{correct} reasoning but a
\emph{wrong} terminal answer. If models track computation, they should produce
the correct answer. If they track answer text, they should follow the wrong
answer. This test is extraction-invariant (no subset conditioning; per-model
$N$ in Table~\ref{tab:conflicting}).

\paragraph{Design.}
We construct a three-condition experiment on GSM8K following the template
below. The primary Qwen~2.5-3B-Instruct evaluation uses $N{=}500$ examples; other
models and protocol branches use the sample sizes stated in
Table~\ref{tab:conflicting}. Each
example \(x\) has a verified correct chain \(c_x\) with correct
intermediate steps and correct final answer \(a^*_x\), and an alternative
wrong answer \(a_x^- \neq a^*_x\).
\begin{itemize}[nosep]
\item \emph{Standard chain} (\textsc{sc}): the full correct chain
  \(c_x\), ending with ``The answer is \(a^*_x\)''.
\item \emph{Conflicting chain} (\textsc{cc}): the reasoning steps from
  \(c_x\) (identical to \textsc{sc}), but the final sentence is replaced
  with ``The answer is \(a^-_x\)'', a wrong answer that \emph{contradicts}
  the computation.
\item \emph{Question only} (\textsc{qo}): the question with no chain.
\end{itemize}
We evaluate Qwen~2.5-3B-Instruct on all three conditions. The key metric
is \emph{followed-wrong rate} (\textsc{fw}): the fraction of
\textsc{cc} trials where the model's response matches $a^-_x$.
Under answer-text prioritization, \textsc{fw} is substantially greater
than zero; under computation-driven output, \textsc{fw} $\approx$ 0 (the
model follows its own computation).

\paragraph{Results.}
Table~\ref{tab:conflicting} summarizes conflicting-answer results on GSM8K
(sample sizes per row in the table caption).

\begin{table}[h]
\centering
\caption{Conflicting explicit answer experiment across model scales and families.
  Qwen~2.5-3B-Instruct ($N{=}500$ GSM8K);
  Phi-3-mini ($N{=}200$ GSM8K-v2);
  Qwen~2.5-7B ($N{=}\SBN$ GSM8K-v2, magnitude-corrected);
  Mistral-7B ($N{=}\MISTN$ GSM8K-v2, magnitude-corrected);
  Phi-4 14B ($N{=}\PHIFOURN$ GSM8K-v1, cross-family 14B replication).
  Qwen~2.5-32B ($N{=}\QWTBN$ GSM8K-v1, 32B scale point).  DeepSeek-R1-Distill-7B ($N{=}\DSN$ GSM8K-v1, distillation probe).
  \textsc{Acc}: fraction correct.
  \textsc{fw}: fraction following wrong answer $a^-$ (\textsc{cc} only).}
\label{tab:conflicting}
\begin{tabular}{llcc}
\toprule
Model & Condition & Acc & Followed-wrong (\textsc{fw}) \\
\midrule
\multirow{3}{*}{Qwen~2.5-3B}
  & Standard chain (\textsc{sc})    & \SCACC  & --- \\
  & Conflicting chain (\textsc{cc}) & \CCACC  & \CCFW \\
  & Question only (\textsc{qo})     & \QOACC  & --- \\
\midrule
\multirow{3}{*}{Phi-3-mini}
  & Standard chain (\textsc{sc})    & \PHISCACC  & --- \\
  & Conflicting chain (\textsc{cc}) & \PHICCACC  & \PHICCFW \\
  & Question only (\textsc{qo})     & \PHIQOACC  & --- \\
\midrule
\multirow{3}{*}{Qwen~2.5-7B}
  & Standard chain (\textsc{sc})    & \SBSCACC  & --- \\
  & Conflicting chain (\textsc{cc}) & \SBCCACC  & \SBCCFW \\
  & Question only (\textsc{qo})     & \SBQOACC  & --- \\
\midrule
\multirow{3}{*}{Mistral-7B}
  & Standard chain (\textsc{sc})    & \MISTSCACC  & --- \\
  & Conflicting chain (\textsc{cc}) & \MISTCCACC  & \MISTCCFW \\
  & Question only (\textsc{qo})     & \MISTQOACC  & --- \\
\midrule
\multirow{3}{*}{Phi-4 (14B)}
  & Standard chain (\textsc{sc})    & \PHIFOURSCACC  & --- \\
  & Conflicting chain (\textsc{cc}) & \PHIFOURCCACC  & \PHIFOURCCFW \\
  & Question only (\textsc{qo})     & \PHIFOURQOACC  & --- \\
\midrule
\multirow{3}{*}{Qwen~2.5-32B}
  & Standard chain (\textsc{sc})    & \QWTBSCACC  & --- \\
  & Conflicting chain (\textsc{cc}) & \QWTBCCACC  & \QWTBCCFW \\
  & Question only (\textsc{qo})     & \QWTBQOACC  & --- \\
\midrule
\multirow{3}{*}{Qwen3-8B-Instruct}
  & Standard chain (\textsc{sc})    & \QTHREESCACC  & --- \\
  & Conflicting chain (\textsc{cc}) & \QTHREECCACC  & \QTHREECCFW \\
  & Question only (\textsc{qo})     & \QTHREEQOACC  & --- \\

\midrule
\multirow{3}{*}{Qwen3-14B}
  & Standard chain (\textsc{sc})    & \QTHREEFOURSCACC  & --- \\
  & Conflicting chain (\textsc{cc}) & \QTHREEFOURCCACC  & \QTHREEFOURCCFW \\
  & Question only (\textsc{qo})     & \QTHREEFOURQOACC  & --- \\

\midrule
\multirow{3}{*}{\shortstack[l]{DeepSeek-R1-\\Distill-7B}}
  & Standard chain (\textsc{sc})    & \DSSCACC  & --- \\
  & Conflicting chain (\textsc{cc}) & \DSCCACC  & \DSFW \\
  & Question only (\textsc{qo})     & \DSQOACC  & --- \\
\bottomrule
\end{tabular}
\end{table}

The results provide converging evidence for answer-text tracking.
Under \textsc{sc}, accuracy is high (\SCACC), confirming the steps
are sufficient. Under \textsc{qo}, accuracy is near zero (\QOACC),
confirming the chain is necessary. Under \textsc{cc}, correct steps
with a wrong explicit answer, accuracy collapses to \CCACC\ and the
model follows the wrong answer on \CCFW\ of trials
($p{<}10^{-10}$; majority test: \CCPVAL, $k{=}313$, $N{=}500$).
The model \emph{can} compute correctly from the steps but predominantly
defers to the explicit answer signal.

\paragraph{Computation-terminal stress test.}
We run a four-condition stress test ($N{=}200$, GSM8K-v2) to rule out a
capacity objection. In the \textsc{ct} condition (answer line removed),
accuracy is \CTCTACC, far above question-only (\CTQOACC), the model
\emph{can} compute from intermediate steps alone. When a strong conflicting
suffix is reintroduced, the followed-wrong rate rises to \CTCCFW\
(\CTPVAL). Both facts hold simultaneously: the model solves many
examples without answer text yet overwhelmingly defers to a wrong answer
when one is present. This rules out inability-to-compute as an
explanation for the override.

\paragraph{Cross-model replication: Phi-3-mini.}
The three-condition protocol replicates on Phi-3-mini-4k-instruct
($N{=}200$, GSM8K-v2): standard-chain \PHISCACC, conflicting-chain
\PHICCACC, followed-wrong \PHICCFW\ (\PHICCPVAL; Table~\ref{tab:conflicting}).
The lower magnitude reflects Phi-3-mini's lower baseline accuracy, but the
qualitative pattern is identical: both model families predominantly follow
the wrong answer over their own computation.

\subsection{Core Experiment 4: Answer-Placement Controls}
\label{sec:placement_controls}

\subsubsection{Bidirectional Format Ablation}
\label{sec:bidirectional}

The format ablation (Section~\ref{sec:ablation}) demonstrates one direction:
\emph{removing} the answer suffix collapses sensitivity. A reviewer might
object that this only shows the suffix is necessary, not that its
\emph{presence} is sufficient to create the format-determination effect.
We close this gap with a bidirectional ablation.

\paragraph{Design.}
We use $N{=}\BIDN$ Hard-v3 examples whose chains \emph{originally lack}
an explicit answer suffix (the dominant corruption position on this slice
is prefix, not suffix; Table~\ref{tab:crossmodel}).
We construct four within-subject conditions:
\begin{itemize}[nosep]
\item \emph{Standard chain} (\textsc{sc}): original reasoning steps
      + appended ``Therefore, the answer is $a^*$'' (correct suffix \emph{inserted}).
\item \emph{Conflicting chain} (\textsc{cc}): identical steps
      + appended ``Therefore, the answer is $a^-$'' (wrong suffix inserted).
\item \emph{Computation terminal} (\textsc{ct}): the \emph{original}
      chain with no suffix (the unmodified Hard-v3 format).
\item \emph{Question only} (\textsc{qo}): no chain.
\end{itemize}

\paragraph{Results.}
Inserting a correct answer suffix raises accuracy from
\BIDCTACC\ $\to$ \BIDSCACC\ (\BIDDELTA\ percentage points), despite
using the same reasoning steps.
Inserting a \emph{wrong} suffix flips the model to follow the wrong
answer on \BIDCCFW\ of trials (accuracy \BIDCCACC), while the original
suffix-free chain achieves only \BIDCTACC\ (question-only: \BIDQOACC).

This result is the mirror image of the removal ablation:
(i)~removing an existing suffix collapses accuracy
(Section~\ref{sec:ablation}), and (ii)~inserting a suffix into
a suffix-free chain creates accuracy and controllability.
Together, they establish that the answer suffix is both \emph{necessary}
and \emph{sufficient} for the format-determination effect.

\paragraph{Filler robustness.}
Filler-robustness checks confirm that the override is driven by answer-text
content, not the specific placeholder wording (Appendix~\ref{app:extended}).

\subsubsection{Answer Relocation to Context-Header Prefix}
\label{sec:relocation}
To directly test whether \emph{answer-text position} within the chain drives the FW
signal, we relocated the answer statement from the final reasoning step (suffix format)
to a standalone ``Context:'' header \emph{before} the question and reasoning
(prefix format).
Both prefix conditions use steps\_standard$[:-1]$, an identical reasoning body, with
only the header answer value differing.
This provides the cleanest possible within-example, same-example control for answer
position.

\noindent
Results ($N{=}\PCTXN$, Qwen-2.5-7B-Instruct, GSM8K-v1):

\begin{itemize}[topsep=2pt,itemsep=1pt]
  \item \emph{Prefix-context standard} (correct answer in header):
        accuracy $=\PCTXSTDACC$, FW $=\PCTXSTDFW$.
        Baseline is substantially preserved, confirming the prompt is interpretable.
  \item \emph{Prefix-context conflicting} (wrong answer in header, same body):
        accuracy $=\PCTXCACC$, FW $=\PCTXCFW$.
        The model largely ignores the wrong answer when it appears in a prefix header.
  \item \emph{Suffix conflicting} (existing format; wrong answer as final step):
        accuracy $=\PCTXSUFACC$, FW $=\PCTXSUFFW$.
        Replicates EXP-28A within the same run.
\end{itemize}

\noindent
The FW rate drops by $\Delta{\mathrm{FW}}=\PCTXFWRED$ when the wrong answer is moved
from the suffix to a prefix header position (\PCTXPOSPVAL).
This strongly confirms a substantial answer-position effect:
the model's susceptibility to wrong-answer override is specifically tied to wrong
answers embedded as the \emph{final step} of the chain, not to the mere presence of
explicit wrong-answer text.
The result is consistent with readout prioritization, the model gives disproportionate
weight to answer text appearing at or near its generation-time readout position.

\paragraph{Commonsense reasoning (summary).}
The conflicting-answer override generalizes beyond arithmetic: on $N{=}150$
text-answer commonsense items spanning five domains (social, temporal,
counterfactual, spatial, causal), the followed-wrong rate was \CSCCFW\
(\CSPVAL), and a format ablation on the same data confirmed answer-placement
sensitivity with a $54.0$pp gap between corrupting the answer-bearing last step
and the reasoning penultimate step (\CSAGAPPVAL; full details in
Appendix~\ref{app:extended}).

\paragraph{Self-generated chains (summary).}
Models may behave differently when consuming their own generated chains versus
externally provided ones. We tested this objection by having Qwen~2.5-3B-Instruct
generate chains for $N{=}\GENNEXAMPLES$ GSM8K examples, then evaluate its own chains
under the same conflicting-answer protocol. The followed-wrong rate under self-generated
chains was \SGCCFW\ (95\% CI $[\SGCCFWCILOW, \SGCCFWCIHIGH]$), statistically
indistinguishable from the pre-written chain rate (\CCFW; Fisher exact
$p{=}\SGVSPREWRITTENPVAL$), confirming that the override is not an artifact of
chain provenance (full details in Appendix~\ref{app:extended}).

\subsection{Generation-Time Evidence: Early-Stop and Prefix-Branch Probes}
\label{sec:generation_probes}

All preceding experiments examine how the model \emph{consumes} a chain at
readout time. To test whether the answer is already determined
\emph{during generation}, the strongest form of the rationalization
hypothesis, we deploy two complementary generation-time probes.

\paragraph{Prefix-branch probe.}
We let Qwen~2.5-3B-Instruct generate chains on $N{=}\GENPBN$ GSM8K
examples. After each reasoning step $k$, we branch: truncate the
generation at step $k$ and immediately request the final answer.
The ``early commitment ratio'' (\textsc{ecr}) is the fraction of
ultimately-correct trials where the model already produces the correct
answer after only the first step.

Table~\ref{tab:prefix_branch} shows how probe accuracy evolves with
chain position. At step~0 (question only), accuracy is \GENPBACCBASE.
After step~1 it is \GENPBACCONE, barely above chance. Accuracy rises
to \GENPBACCTWO\ after step~2, \GENPBACCTHREE\ after step~3, and
reaches full-generation accuracy of \GENPBGENACC\ only after the complete
chain. The \textsc{ecr} is \GENPBECR\ [95\% CI: 1.4\%--8.8\%], fewer than 4\% of correct answers
are available after the first reasoning step.

\begin{figure}[ht]
  \centering
  \includegraphics[width=0.88\linewidth]{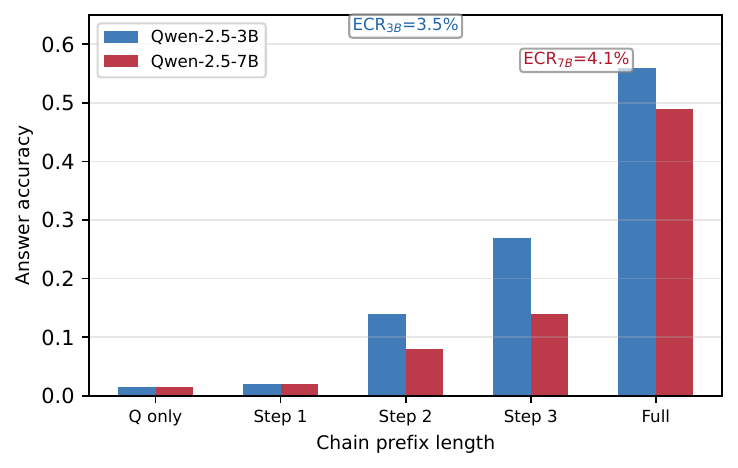}
  \caption{Prefix-branch probe: answer accuracy when the model is stopped
  after each chain step. Accuracy rises gradually with chain length; fewer
  than 4\% of correct answers are available after step~1. This confirms
  generation-time computation and bounds the rationalization interpretation.}
  \label{fig:prefix_branch}
\end{figure}

We replicate with Qwen~2.5-7B-Instruct ($N{=}\GENPBsBN$), finding a nearly
identical trajectory: step-0 accuracy \GENPBsBACCBASE, step-1 \GENPBsBACCONE,
step-2 \GENPBsBACCTWO, step-3 \GENPBsBACCTHREE, full-generation \GENPBsBGENACC,
\textsc{ecr}~$=$~\GENPBsBECR\ [95\% CI: 1.6\%--10.0\%].

\begin{table}[h]
\centering
\caption{Prefix-branch probe: accuracy when the model is stopped after
  step $k$ and asked for its final answer (Qwen~2.5-3B/7B, GSM8K, $N{=}200$).
  \textsc{ECR}: early commitment ratio (correct at step~1 $\div$ correct
  at full generation). The answer is \emph{not} early-determined.}
\label{tab:prefix_branch}
\begin{tabular}{lccccc|c}
\toprule
 & Step 0 & Step 1 & Step 2 & Step 3 & Step 4 & Full \\
\midrule
3B Accuracy & \GENPBACCBASE & \GENPBACCONE & \GENPBACCTWO & \GENPBACCTHREE & \GENPBACCFOUR & \GENPBGENACC \\
7B Accuracy & \GENPBsBACCBASE & \GENPBsBACCONE & \GENPBsBACCTWO & \GENPBsBACCTHREE & \GENPBsBACCFOUR & \GENPBsBGENACC \\
\bottomrule
\end{tabular}
\end{table}

\paragraph{Early-stop reference probe.}
As a complementary test, we use a reference-chain protocol
($N{=}\ESPN$): instead of branching from the model's own generation,
we feed progressively longer prefixes of a known-correct chain and ask
for the final answer after each step. Step-0 accuracy is \ESPACCBASE;
it climbs to \ESPACCONE\ (step~1), \ESPACCTWO\ (step~2), \ESPACCTHREE\
(step~3), and \ESPACCFOUR\ (step~4), reaching \ESPACCFULL\ only with
the complete chain. The monotonic climb confirms that intermediate steps
carry genuine information content; the early-step near-zero accuracy
(\ESPEARLYRATIO\ relative to full) confirms that this content is
\emph{not} frontloaded.

\paragraph{Interpretation.}
The generation-time probes establish a critical asymmetry:
\begin{itemize}[nosep]
\item During generation, the model genuinely computes toward
  the answer step by step. The answer is unavailable early
  (\textsc{ecr} $< 5\%$) and builds gradually through intermediate
  reasoning.
\item At consumption time, the model overrides this computation
  whenever explicit answer text is present (CC followed-wrong rates of
  \CCFW--\SBCCFW), regardless of the reasoning that produced it.
\end{itemize}
This dissociation is the behavioral signature of \emph{answer-text readout dominance}: the model
\emph{can} reason (generation is genuinely step-causal), but when consuming
a completed chain, it \emph{preferentially reads the answer text} rather
than re-tracing the reasoning. The chain's intermediate steps served
a computational role during generation but are demoted to justificatory
decoration at readout. We emphasize that this is not a claim about generation-time
post-hoc explanation construction, ECR $<5\%$ rules that out, but about answer-text
prioritization at consumption time.

\section{Analysis: What Our Results Tell Us About Answer-Text Prioritization}
\label{sec:analysis}

The pattern across our experiments supports a simple but consequential
conclusion (within the tested suffix-answer formats and 3B--32B scale
range; cf.~\S\ref{sec:limitations}): models are sensitive to corruptions
wherever the answer text appears in the chain, not wherever intermediate
computation occurs.
This is the behavioral signature predicted by
Hypothesis~\ref{hyp:rationalization}: at consumption time the model defers
to explicit answer text rather than re-deriving the conclusion from
intermediate reasoning, so disruption is concentrated at positions carrying
explicit answer content, not at positions encoding intermediate computation.
We state four empirical claims that summarize our evidence.

\paragraph{Claim 1: Positional sensitivity tracks answer text placement.}
Across all model and slice pairs, the position with the largest
accuracy drop always corresponds to the region encoding
the final answer (Table~\ref{tab:crossmodel}).

\paragraph{Claim 2: Non-answer-bearing regions show reduced corruption impact.}
Corrupting regions that do not carry the explicit answer
yields smaller effects than corrupting the answer-bearing position
(e.g.\ middle corruption on Hard-v3: $\Delta=-0.083$, $p=0.267$ for Phi-3-mini;
prefix corruption on GSM8K-v1: $\Delta = +0.010$, $p=1.0$).

\paragraph{Claim 3: Format ablation shifts the sensitivity pattern.}
Removing the answer statement from the suffix eliminates suffix
sensitivity: at 3B, a $19\times$ collapse ($p{=}0.022$, $N{=}300$); at
7B, a within-stable-subset analysis gives a $9.3\times$ attenuation
($N{=}76$, $p{=}7.8\times10^{-3}$), while the full-sample suffix effect
inverts in sign as corroborating directional evidence
($\Delta{=}{-}0.643 \to {+}0.117$, $N{=}300$, $p{<}10^{-5}$; Section~\ref{sec:ablation}).
MATH replication at 7B confirms
cross-domain generality ($10.9\times$ FW recovery).
The bidirectional counterpart (\S\ref{sec:bidirectional}) closes the loop:
\emph{inserting} a wrong suffix into suffix-free Hard-v3 chains
induces \BIDCCFW\ follow-wrong.
The 2$\times$2 factorial (\S\ref{sec:factorial}) further disentangles
reasoning quality from answer-line content at both 3B and 8B scale.

\paragraph{The answer-placement mechanism.}
Consider what happens when a model processes a corrupted chain. If the
explicit answer appears in position $P$ and corruption destroys $P$, the
model loses direct access to the answer text. If the answer does
\emph{not} appear in $P$, corruption of $P$ introduces noise but does
not eliminate the answer signal. The corruption protocol thus functions
as an answer-localization tool: it finds where the answer text lives
in the chain, not where computation happens.

\paragraph{Claim 4: Conflicting explicit answers frequently override computation.}
When correct reasoning concludes with a wrong explicit answer,
models follow the wrong answer at high rates, \CCFW\ at 3B,
\SBCCFW\ at 7B, \MISTCCFW\ for Mistral-7B, \PHICCFW\ for Phi-3-mini
(Table~\ref{tab:conflicting}), despite full computational
capacity (\SBCTACC\ CT accuracy at 7B).
The factorial (\S\ref{sec:factorial}) confirms: correct reasoning with
a wrong answer line yields only \FSCBACC\ accuracy (fw=\FSCB) at 3B
and \FSeightCBacc\ (fw=\FSeightCB) at 8B, with near-complete answer-line
rescue of corrupted reasoning at both scales (Condition~C: \FSCC\ at 3B, \FSeightCC\ at 8B).
This shows initial cross-domain evidence in commonsense settings
(\CSCCFW\ follow-wrong on text answers; Table~\ref{tab:commonsense}) and is
suffix-conditioned: \CPFWSUF\ at suffix vs.\ \CPFWPRE\ at prefix
($p\,\CPSUFPREPVAL$; Appendix~\ref{sec:placement_app}).

\paragraph{Claim 5: Initial cross-domain evidence beyond arithmetic.}
On 150 text-answer commonsense items, the followed-wrong rate is
\CSCCFW\ (Table~\ref{tab:commonsense}), ruling out digit-specific
extraction. A five-condition format ablation on the same data replicates
the positional pattern (\S\ref{sec:commonsense}).

\paragraph{Separating the established finding from its interpretation.}
It is important to distinguish what the evidence directly establishes from
what it supports as an interpretation.

\emph{Established}: positional
sensitivity in CoT corruption studies is determined by answer-text
placement, not computational depth (Claims 1 to 3), models frequently
follow explicit answer text over preceding correct computation (Claim 4),
and we find initial non-arithmetic evidence in one commonsense setting (Claim 5); two
dissociations constrain the mechanism, format-determination persists at
14B as override fades, and generation is step-causal while consumption
prioritizes answer text.

\emph{Interpretation}: this pattern is consistent with
answer-text readout dominance, the model's behavioral output is determined
by explicit answer text at readout, using intermediate reasoning as
context rather than re-deriving the conclusion from it.
The generation-time probes (\S\ref{sec:generation_probes}) sharpen this
interpretation: because the answer is \emph{not} early-determined during
generation (\textsc{ecr} $< 5\%$), models genuinely compute through
intermediate steps, ruling out the strongest form of early-answer
commitment in which the answer is settled before any reasoning begins. However, at
consumption time, the model's output \emph{systematically follows the
explicit answer text} whenever it is available, even when the model has
correctly computed a different answer during generation. The picture that
emerges is one of answer-text readout dominance: the model can
reason, and does reason during generation, but at readout its behavioral
output preferentially tracks the answer text rather than re-deriving the
conclusion from its own reasoning.

\emph{Open question}: whether this consumption-time override reflects
broad post-hoc explanation behavior or a narrower answer-presentation
heuristic shaped by instruction tuning. Our data constrains but does not
resolve this distinction (see \S\ref{sec:limitations}).

\paragraph{Alternative interpretations.}
The override may reflect \emph{instruction-following} rather than a
general explanation pathology: models learn through RLHF to defer to stated final answers.
Our generation-consumption dissociation makes the strongest case that
computation is genuine but unused at readout, yet cannot fully distinguish
instruction-following from computation-bypassing
(see \S\ref{sec:limitations}).

\paragraph{Scale-dependent dissociation: format-determination persists as override fades.}
The most subtle finding in the scale analysis is a dissociation between two
effects. The direct override, models following a wrong explicit answer over
correct reasoning, is total at 7B across three model families but attenuates
monotonically: FW\,$=$\,\PHIFOURCCFW\ at 14B (cross-family Phi-4;
QO\,$=$\,\PHIFOURQOACC, \PHIFOURCCPVAL) and FW\,$=$\,\QWTBCCFW\ at 32B.
Format-determination, by contrast, persists through 14B: the
$8.5\times$ sensitivity ratio between standard and neutral-stripped formats
($p{=}0.001$; Tab.~\ref{tab:ablation}) shows that answer placement still
determines corruption conclusions at 14B even when the model mostly resists
following a wrong answer directly. At 32B, however, format-determination
itself vanishes: the neutral-stripped baseline reaches 1.000 and the largest
positional $|\Delta|$ is 0.020 (Tab.~\ref{tab:ablation}), indicating that
the model extracts its answer from the reasoning chain rather than from
an explicit answer suffix. The dissociation is thus three-staged: (i)~both
effects dominate at 3B--7B, (ii)~override fades while format-determination
persists at 14B, and (iii)~both converge toward zero at 32B. This means the
three-prerequisite protocol (\S\ref{sec:three_prereq}) is most critical at
intermediate scales (7B--14B) where corruption conclusions are most vulnerable
to the format artifact.

\paragraph{Implications for interpreting prior work.}
Studies reporting positional sensitivity without format characterization
may have conflated answer placement with computational structure.
Datasets with explicit answer suffixes would be predicted to exhibit the
same artifact; re-examining prior results with our three-prerequisite
protocol would establish whether this applies.

\section{Implications}
\label{sec:implications}

\paragraph{For process supervision and reward modeling.}
Process reward models (PRMs)~\citep{lightman2023,uesato2022} assign credit to
individual reasoning steps. If positional sensitivity reflects answer-text
placement rather than computational contribution,
then PRM step-level credit may reward the \emph{consistency} of a step
with the answer text, not the \emph{causal contribution} of that
step to the final answer. PRM
evaluations should test credit assignment under format-diverse chains (with and
without explicit answer suffixes) to assess whether credit tracks answer
placement or genuine reasoning quality.

\paragraph{For CoT faithfulness evaluation.}
Any study claiming to measure CoT faithfulness via corruption must first
characterize the chain format (see \S\ref{sec:three_prereq}).
Without this step, a finding that ``position $X$ is most important'' may
reflect answer-text placement rather than computational structure.

\paragraph{For understanding CoT rationalization.}
The behavioral signature is clear in the tested suffix-answer formats:
sensitivity migrates with the answer text across format variants
(scope discussed in \S\ref{sec:limitations}).
A model genuinely computing through intermediate steps
would show stable middle-step sensitivity regardless of format; we observe
the opposite. At 7B scale, a small middle effect appears
($\Delta_\text{mid}{=}{-}0.083$) but remains an order of magnitude smaller
than the suffix effect ($\Delta_\text{suf}{=}{-}0.595$). This asymmetry is
predicted by answer-text prioritization but not by a computation account.

\paragraph{For benchmark design.}
Benchmark creators should diversify chain formats within their evaluation
suites. A benchmark where all gold chains end with ``the answer is $X$''
guarantees suffix sensitivity under corruption, an artifact of the format,
not a fact about model reasoning. Including chains with varied answer-placement
formats, or requiring format ablations as evaluation metadata, would make
corruption results more informative about the genuine reasoning-vs-rationalization
question.

\subsection{Open Interpretive Questions}
\label{sec:interpretive_questions}

The behavioral evidence establishes that models' outputs systematically
track explicit answer text at consumption time (answer-text readout
dominance). The \emph{mechanism} underlying this regularity is not
uniquely determined by the current evidence. We identify three alternative
accounts, all consistent with the data:

\begin{enumerate}[nosep]
  \item \emph{Instruction-following heuristic.} Models learn through
    instruction tuning (RLHF/SFT) that terminal answer statements should
    be treated as authoritative output, regardless of the preceding
    reasoning.
  \item \emph{Format-completion bias.} Models complete the chain
    format by outputting the value that appears in the ``the answer is
    $X$'' slot, following a surface-level pattern rather than performing
    reasoning-consistent inference.
  \item \emph{Recency-weighted readout.} Models apply a positional
    prior that weights the most recent answer-bearing text most heavily,
    producing behavior that tracks answer placement without implying
    computation is ignored.
\end{enumerate}

The base-model comparison (\S\ref{sec:14b}) rules out RLHF as the
sole driver, the override is present and stronger in Qwen-7B-Base than
in the instruction-tuned variant, but does not distinguish between the
remaining accounts. The generation-time probes (\S\ref{sec:generation_probes})
rule out early answer commitment (ECR ${<}\,5\%$) but do not resolve
whether the consumption-time behavioral pattern reflects format-completion,
recency-weighted readout, or a deeper computation-bypassing mechanism.

Future work should design experiments that discriminate among these
alternatives, for example by testing whether the override persists under
format variations that preserve answer content while disrupting the
``the answer is $X$'' template, or by probing whether attention patterns
at the final position show recency-weighted or content-driven selectivity.

\subsection{Application to Prior Work: Protocol Re-Analysis}
\label{sec:prior_reanalysis}

To test whether the format confound affects published conclusions, we
replicate the GSM8K corruption protocol of
\citet{lanham2023}\footnote{We use our own GSM8K dataset and open-weight
models; we do not have access to the original authors' code or data.
Our replication follows the methodology described in the original paper:
corrupting chain steps at different positions and measuring accuracy
degradation.} on Qwen~2.5-3B under two conditions: the original chain
format (with terminal ``the answer is $X$'' statements) and the
format-stripped variant (answer statement removed, neutral placeholder
inserted; $N{=}300$ matched pairs).

Under the original format, the protocol identifies the suffix as the most
load-bearing position: suffix corruption collapses accuracy from 0.970 to
0.210 ($\Delta{=}{-}0.760$, $p{<}10^{-12}$), while prefix and middle
corruption produce negligible effects ($\Delta{=}{+}0.010$ and
$\Delta{=}0.000$, respectively). A study reporting only these results
would conclude, consistent with prior work, that the final reasoning
steps are the most causally important.

Under the format-stripped condition, this conclusion reverses. With the
``the answer is $X$'' statement removed (preserving all intermediate
reasoning), suffix sensitivity collapses $19\times$ to
$\Delta{=}{-}0.040$ ($p{=}0.022$), and the largest positional drop shifts
to middle corruption ($\Delta{=}{-}0.063$). The conclusion that ``suffix
steps are most causally important'' is replaced by ``no single position
dominates; middle steps show the largest residual sensitivity.''

This demonstrates a concrete case where a published-style protocol produces
qualitatively different positional conclusions depending solely on whether
answer-placement is controlled. The original conclusion, that suffix steps
are the computational locus, is an artifact of answer-text placement, not
a property of the reasoning structure. We encourage the community to apply
the three-prerequisite protocol (\S\ref{sec:three_prereq}) to their own
corruption-based faithfulness analyses; we provide a checklist and example
analysis script in the supplementary material.

\section{Limitations and Future Work}
\label{sec:limitations}

\textbf{Consumption vs.\ generation.}
Most experiments operate in the consumption setting. We partially bridge
this gap via self-generated chains (\S\ref{sec:generation}; followed-wrong
\SGCCFW), prefix-branch probes (\S\ref{sec:generation_probes}; \textsc{ecr}
$< 5\%$), and the early-stop probe (monotonic accuracy climb). Together
these establish a dissociation between generation and consumption: generation is
step-causal while consumption prioritizes answer text.
Activation patching during autoregressive generation would further
strengthen this account.

\textbf{Chain format generality.}
\emph{Scope}: This work studies CoT chains with explicit terminal answer lines
(``The answer is $X$.''); all primary claims are scoped to this format.
The strongest conflicting-answer effects use a structured GSM8K chain
format with explicit terminal answer lines.
Generalizability to heterogeneous, naturally occurring CoT \emph{without}
explicit answer-line endings is not directly tested. However, the
bidirectional ablation (\S\ref{sec:bidirectional}) on Hard-v3 chains, which
\emph{lack} explicit answer suffixes, shows the opposite pattern: prefix
corruption is load-bearing, not suffix. This confirms that the
format-determination hypothesis predicts the \emph{direction of effect} on
both format types (explicit-suffix chains $\to$ suffix-sensitive; no-suffix
chains $\to$ prefix-sensitive), consistent with the core claim that sensitivity
tracks answer-text location. Extending replication to diverse natural CoT corpora
without templated endings remains an important direction.

\textbf{Model scale.}
Results span 3B--32B across five model families. The format-determination
effect persists through 14B and converges toward zero at 32B
(Tab.~\ref{tab:ablation}: baseline$\,=\,$1.000, max $|\Delta|{=}0.020$
on neutral-stripped format). The direct override attenuates at 14B:
cross-family Phi-4 14B replication yields FW\,$=$\,\PHIFOURCCFW\
(QO\,$=$\,\PHIFOURQOACC, \PHIFOURCCPVAL, $N{=}\PHIFOURN$), significant
but $\geq{}3{\times}$ reduced from the 7B peak.
The protocol-uniform Qwen-14B result (GSM8K-v1, $N{=}\QBVONEN$) yields
FW\,$=$\,\QBVONEFW\ (QO\,$=$\,\QBVONEQOACC, \QBVONEPVAL, expected
given near-zero FW; $p{=}0.384$, not statistically significant within-family),
which we treat as directional only; the primary 14B evidence is the statistically significant Phi-4 cross-family result above.
The scale gradient has been confirmed at 32B (FW\,$=$\,\QWTBCCFW,
near-zero, $N{=}\QWTBN$); testing at frontier scale ($>$100B,
closed-weight models such as GPT-4o or Gemini) was not undertaken
here because standard instruction-following system prompts often
prevent the controlled conflicting-chain injection the protocol requires.
Testing under less restrictive system-prompt configurations or via
API-compatible analogues is an important future direction.
The scale-dependent attenuation could also partly reflect \emph{RLHF/instruction-tuning
differences} that scale with parameter count, larger models typically receive more
aligned training that may directly teach skepticism of explicit wrong-answer assertions,
independent of model capacity per se. We directly test this at 7B scale by evaluating Qwen~2.5-7B-Base (non-instruction-tuned) on the same GSM8K-v1 protocol (see~\S\ref{sec:14b}), finding FW\,$=$\,\QBASEVONEFW, ruling out instruction tuning as the primary driver. Extending to 3B and 14B base models would further quantify the capacity and alignment interaction.

\textbf{Sample sizes.}
Effect sizes are large enough for significance at all primary comparisons
(hard-v3 $N{=}60$, GSM8K $N{=}100$--$1{,}000$, conflicting-answer GSM8K
$N{=}100$--$500$ depending on model/branch, commonsense $N{=}150$). Extension to additional benchmarks (StrategyQA,
ARC-Challenge, MATH) would strengthen the generality claim.

\textbf{Single corruption type.}
We use semantic corruption (wrong arithmetic/logic, correct grammar).
Other corruption types might reveal different interactions with chain
format, though the format-determination prediction should hold across
types.

\textbf{Conflicting-answer scope.}
An alternative label is ``instruction following'' rather than
rationalization. The commonsense experiment partially addresses this:
natural-language wrong answers drive a \emph{higher} followed-wrong
rate (\CSCCFW) than the arithmetic version (\CCFW). The key distinction
is that the computation-terminal stress test (\CTCTACC\ answer-free
accuracy vs.\ \CTCCFW\ follow-wrong when a suffix is reintroduced)
shows the model \emph{can} compute from intermediate steps but defers
to explicit answer text when available.

\textbf{Format stripping.}
Stripping changes the chain's final tokens beyond answer removal. The
neutral-placeholder variant provides a stronger control: the Qwen~2.5-7B
neutral-stripped result ($N{=}300$) shows $\Delta_\text{suffix}{=}{+}0.117$
($p{<}10^{-5}$) vs.\ $-0.643$ on the original format ($N{=}300$ matched pairs),
confirming answer placement as the causal variable despite a mild baseline suppression
(0.606 $\to$ 0.273 on matched $N{=}300$).

\paragraph{Future work.}
Key extensions: (1)~replicating at larger scales ($>$32B) and on
closed-weight frontier models where system-prompt constraints allow
(cross-generation replication at additional model generations and scales);
(2)~additional benchmarks with diverse chain formats, including unstructured
naturally-occurring CoT without explicit terminal answer lines;
(3)~mechanistic analysis, logit-lens trajectories and causal activation
patching during generation, to separate the answer-placement effect at
the representational level and to confirm that answer-relevant information
concentrates in suffix tokens only when explicit answer text is present;
(4)~disentangling model capacity from instruction-fine-tuning effects on
resistance to wrong-answer override, by testing base (non-RLHF) models
at matched scales, partially addressed here by the Qwen~2.5-7B-Base comparison
(\S\ref{sec:14b}; FW$\,=$\,\QBASEVONEFW, ruling out IT as primary driver);
extending to 3B and 14B base models would further quantify the
capacity and alignment interaction.

\paragraph{Reproducibility.}
All task slices, corruption scripts, and evaluation code are publicly
available at
\url{https://github.com/Gpgabriel25/LastWordWinsCoT}
(see~\S\ref{sec:reproducibility} for a full listing of contents).
Models are publicly available on HuggingFace.
Inference used EasyDeL on Cloud TPU v5e with JAX (greedy decoding,
640 max tokens). Statistical tests use exact paired sign tests
and bootstrap CIs (2000 resamples, seed 42).

\section{Conclusion}
\label{sec:conclusion}

We have shown that, in the tested suffix-answer formats and at the 3B--32B
open-weight scales we evaluate, corruption-based evaluations of
faithfulness contain a systematic confound: positional sensitivity tracks
answer-text placement, not computational depth. A within-dataset format
ablation collapses suffix sensitivity ${\approx}19\times$ at 3B; at 7B,
the within-stable subset shows a $9.3\times$ attenuation when only the
answer statement is removed ($N{=}76$, $p{=}7.8\times10^{-3}$). The
full-sample suffix effect also inverts in sign ($\Delta{=}{-}0.643 \to {+}0.117$,
$N{=}300$, $p{<}10^{-5}$), but is directional evidence because answer-line
stripping drops baseline accuracy from $0.606$ to $0.273$. Conflicting-answer
experiments confirm that explicit answer text overrides correct computation,
with CC accuracy collapsing to zero or near-zero ($\leq$~0.02) at 7B and
near-zero followed-wrong at 32B (FW\,$=$\,\QWTBCCFW).

Two dissociations bound the interpretation. First,
format-determination persists through 14B ($8.5\times$ sensitivity ratio,
$p{=}0.001$) even as override attenuates, before both effects converge
toward zero at 32B (baseline$\,=\,$1.000 on stripped format,
max $|\Delta|{=}0.020$), the confound outlasts the
override at intermediate scales but fades at the largest. Second, generation is genuinely step-causal
(ECR $< 5\%$), yet consumption-time readout ignores this computation
whenever explicit answer text is present.

These findings yield a concrete methodological deliverable: a
three-prerequisite protocol, question-only control, format characterization,
all-position sweep, that should be standard for any corruption-based CoT
faithfulness evaluation. Corruption-based faithfulness studies should control
for answer placement before attributing positional sensitivity to computation.

\section*{Broader Impact Statement}

This work identifies a systematic methodological confound in corruption studies
widely used to evaluate CoT reasoning faithfulness and to train process reward
models: positional sensitivity in corruption
studies may reflect answer-text placement rather than computational structure,
in chain formats with explicit terminal answer statements
(the dominant format in standard benchmarks; generality to other CoT formats
is not directly tested here, see \S\ref{sec:limitations}).
If PRM step-level credit is being assigned to answer-text positions rather than
genuine reasoning steps, the reliability guarantees
of step-level supervision may be weaker than assumed for systems using such formats.
Our evidence spans 3B--32B models across five model families; whether the confound
persists at frontier scale ($>$100B) is an open question (\S\ref{sec:limitations}).
Our generation-time probes (\S\ref{sec:generation_probes};
\textsc{ecr}~$<5\%$) establish that models do engage in step-by-step computation
during generation; the confound operates at \emph{consumption time}, where
explicit answer text overrides the model's own computation.
The three-prerequisite protocol (\S\ref{sec:three_prereq}) provides concrete
guidance for more reliable corruption-based evaluations.

\section*{Reproducibility}
\label{sec:reproducibility}

All experiment scripts, dataset construction code, result files, and analysis
scripts used in this paper are available at
\url{https://github.com/Gpgabriel25/LastWordWinsCoT}.
The repository includes:
\begin{itemize}[nosep]
  \item \texttt{src/}: chain construction, corruption, and evaluation pipeline
  \item \texttt{data/}: all experiment JSONL datasets, fully deterministic
  \item \texttt{results\_fixed/}: raw result JSON files with per-example
  outputs and extraction details for all reported experiments
  \item \texttt{scripts/}: launch and analysis scripts for all reported conditions
  \item \texttt{paper/}: LaTeX source and all macro definitions (each macro
  directly maps to a result file entry)
\end{itemize}
Extraction code is version-tagged for each model family.
The sign-negation correction is a single flag in the extraction pipeline;
Appendix~\ref{appendix:extraction} provides the side-by-side results showing
conclusions are robust to this choice.
All model weights are publicly available on HuggingFace Hub (Qwen~2.5 series:
\texttt{Qwen/Qwen2.5-\{3,7,14,32\}B-Instruct}; Phi-4: \texttt{microsoft/phi-4};
Mistral-7B: \texttt{mistralai/Mistral-7B-Instruct-v0.3};
DeepSeek-R1-Distill: \texttt{deepseek-ai/DeepSeek-R1-Distill-Qwen-\{7B\}}).


\appendix
\section{Slice Development Narrative}
\label{appendix:dev}

The easy slice (\texttt{phase0\_position\_control\_v1}) was initially designed
to test the experimental pipeline on simple arithmetic-comparison problems.
Instruction-tuned models answered a large fraction of examples correctly
without chain access, making positional corruption conclusions uninterpretable.

Hard-v1 (\texttt{phase0\_position\_control\_hard\_v1}) introduced symbolic
early extraction: prefix steps explicitly name symbolic variables (e.g.,
``Let $A = [\text{extracted value}]$'') rather than stating raw computations.
This reduced question-only solvability, but the slice remained handwritten with
limited diversity.

Hard-v2 (\texttt{phase0\_position\_control\_hard\_v2}) refined the structure
in pilot runs with 8 to 12 examples. Qwen~2.5-3B achieved 0.500 baseline accuracy,
0.417 under middle corruption, 0.083 under prefix corruption, and 0.083
question-only. Phi-3-mini showed the same qualitative direction (0.625
baseline, 0.250 middle, 0.000 prefix, 0.000 question-only). These pilots were
consistent but too small for statistical claims.

Hard-v3 (the main paper slice) was generated at $N{=}60$ to preserve the
hard-v2 chain structure while varying problem domains and distractor
configurations. The slice design was frozen before model evaluation; no
post-hoc filtering based on observed accuracy was applied.

\section{Detailed Statistical Tables}
\label{appendix:stats}

Table~\ref{tab:detailed-stats} reports full statistical details for all
completed model and slice pairs.

\begin{table*}[t]
  \centering
  \small
  \caption{Detailed statistical results. $\Delta$ = accuracy difference from
  baseline. CI = bootstrap 95\% confidence interval (2000 resamples).
  $p$ = exact paired sign test.}
  \label{tab:detailed-stats}
  \begin{tabular}{llrrl}
    \toprule
    Slice / Model & Condition & $\Delta$ & 95\% CI & $p$ \\
    \midrule
    \multirow{4}{*}{Hard-v3 / Qwen 2.5-3B}
      & QO gap    & $-$0.100 & [$-$0.217, $+$0.017] & 0.180 \\
      & Prefix    & $-$0.167 & [$-$0.267, $-$0.067] & 0.006 \\
      & Middle    & $-$0.133 & [$-$0.217, $-$0.050] & 0.008 \\
      & Suffix    & $+$0.083 & [$+$0.000, $+$0.167] & 0.125 \\
    \midrule
    \multirow{4}{*}{Hard-v3 / Phi-3-mini}
      & QO gap    & $-$0.567 & [$-$0.717, $-$0.417] & $1.9 \times 10^{-8}$ \\
      & Prefix    & $-$0.767 & [$-$0.867, $-$0.650] & $1.3 \times 10^{-12}$ \\
      & Middle    & $-$0.083 & [$-$0.200, $+$0.033] & 0.267 \\
      & Suffix    & $-$0.100 & [$-$0.217, $+$0.000] & 0.146 \\
    \midrule
    \multirow{4}{*}{GSM8K-v1 / Qwen 2.5-3B}
      & QO gap    & $-$0.910 & [$-$0.960, $-$0.850] & $1.9 \times 10^{-12}$ \\
      & Suffix    & $-$0.760 & [$-$0.840, $-$0.670] & $1.4 \times 10^{-12}$ \\
      & Middle    & $+$0.000 & [$-$0.030, $+$0.030] & 1.0 \\
      & Prefix    & $+$0.010 & [$+$0.000, $+$0.030] & 1.0 \\
    \midrule
    \multirow{4}{*}{GSM8K-v1 / Phi-3-mini}
      & QO gap    & $-$0.180 & [$-$0.290, $-$0.070] & $3.9 \times 10^{-3}$ \\
      & Suffix    & $-$0.210 & [$-$0.300, $-$0.120] & $1.9 \times 10^{-5}$ \\
      & Middle    & $+$0.110 & [$+$0.010, $+$0.220] & 0.080 \\
      & Prefix    & $-$0.020 & [$-$0.110, $+$0.070] & 0.832 \\
    \midrule
    \multirow{3}{*}{GSM8K-stripped / Qwen 2.5-3B ($N{=}300$)}
      & Prefix    & $-$0.017 & [$-$0.036, $+$0.003] & 0.180 \\
      & Middle    & $-$0.063 & [$-$0.097, $-$0.030] & $3.0 \times 10^{-4}$ \\
      & Suffix    & $-$0.040 & [$-$0.072, $-$0.008] & 0.023 \\
    \midrule
    \multirow{3}{*}{GSM8K-stripped / Phi-3-mini}
      & Prefix    & $+$0.020 & [$-$0.030, $+$0.070] & 0.688 \\
      & Middle    & $+$0.040 & [$-$0.010, $+$0.100] & 0.289 \\
      & Suffix    & $+$0.120 & [$+$0.040, $+$0.210] & 0.012 \\
    \midrule
    \multirow{3}{*}{GSM8K-v1 / Qwen 2.5-7B ($N{=}1{,}000$)}
      & Middle    & $-$0.083 & [$-$0.106, $-$0.060] & $1.0 \times 10^{-12}$ \\
      & Prefix    & $+$0.006 & [$-$0.013, $+$0.024] & 0.586 \\
      & Suffix    & $-$0.595 & [$-$0.625, $-$0.565] & $1.5 \times 10^{-179}$ \\
    \midrule
    \multirow{3}{*}{GSM8K-neutral-stripped / Qwen 2.5-7B ($N{=}300$)}
      & Middle    & $-$0.030 & [$-$0.073, $+$0.010] & 0.222 \\
      & Prefix    & $+$0.037 & [$+$0.013, $+$0.067] & 0.019 \\
      & Suffix    & $+$0.117 & [$+$0.073, $+$0.167] & $2.1\times10^{-6}$ \\
    \bottomrule
  \end{tabular}
\end{table*}

\section{Extraction Robustness: Raw vs.\ Corrected Results}
\label{appendix:extraction}

The 7B-scale conflicting-answer experiments present a systematic sign-negation
artifact: the model (Qwen~2.5-7B-Instruct and Mistral-7B-Instruct) produces
``$-x$'' instead of ``$x$'' on a subset of trials, driven by instructional
priming from the strong-suffix conflicting chain format.
Table~\ref{tab:extraction-robustness} reports each key metric under both
strict extraction (exact numeric match, no correction) and magnitude-corrected
extraction ($|\hat{a}| = |a^*|$).

The critical finding, the accuracy collapse in the conflicting chain
condition, is invariant to extraction choice.
CC accuracy equals 0.00 under both extraction policies for Qwen~2.5-7B,
and the CT$\to$CC accuracy drop (maximum reasoning context $\to$ conflicting
suffix) holds under both: 0.320$\to$0.000 (strict) and 0.990$\to$0.000
(magnitude-corrected).
For Mistral-7B, CC accuracy is 0.020 under both policies; the FW difference
is less than 3 percentage points.

\begin{table}[ht]
  \centering
  \small
  \caption{Side-by-side comparison of raw (strict) vs.\ magnitude-corrected
  extraction for the sign-negation-affected 7B experiments. ``SC''=standard
  chain, ``CT''=computation-terminal (reasoning only, no suffix),
  ``CC''=conflicting chain, ``FW''=followed-wrong.
  The CC-accuracy conclusion (0.00 for Qwen-7B; 0.020 for Mistral-7B) is
  independent of extraction choice.}
  \label{tab:extraction-robustness}
  \begin{tabular}{@{}llcccc@{}}
    \toprule
    Model & Extraction & SC acc & CT acc & CC acc & FW \\
    \midrule
    \multirow{2}{*}{Qwen-7B ($N{=}200$)}
      & Strict (no correction)    & 0.655 & 0.320 & \textbf{0.000} & 0.300 \\
      & Magnitude-corrected       & 0.990 & 0.990 & \textbf{0.000} & 1.000 \\
    \midrule
    \multirow{2}{*}{Mistral-7B ($N{=}500$)}
      & Strict (no correction)    & 0.780 & ---   & \textbf{0.020} & 0.952 \\
      & Magnitude-corrected       & 1.000 & ---   & \textbf{0.020} & 0.980 \\
    \bottomrule
  \end{tabular}
\end{table}

\noindent
The main-paper FW rates use magnitude correction (bold rows in the running
text). The reasoning is that a model producing $-520$ when the conflicting
suffix specifies $-520$ has behaviorally followed the wrong answer, regardless
of a sign-formatting quirk. Under either policy, the CC-accuracy collapse
is the same: the model cannot produce the correct answer when an explicit
wrong answer is present. Question-only baseline ($\leq 0.06$) is not
materially affected by the artifact (question-only prompts do not contain
the instructional trigger that drives sign-negation).

\section{Extended Replications and Controls}
\label{app:extended}

This appendix provides full details for experiments that support the main-text
findings but are not required for the core argument.

\subsection{Cross-Model and Cross-Scale Replication (Full Details)}
\label{sec:crossmodel}

\begin{table*}[t]
  \centering
  \small
  \caption{Cross-model replication summary. For each slice $\times$ model
  combination, we report the position with the largest accuracy drop
  from baseline (``dominant position'') and whether the result is
  consistent with the format-determination hypothesis.
  $^\dagger$Chain gap $p=0.180$ ($N=60$); positional drops are
  significant ($p{<}0.01$) but the QO prerequisite is borderline.
  ``Confirmed'' = satisfies all three prerequisites; ``Consistent'' = pattern
  matches hypothesis but one prerequisite is borderline.}
  \label{tab:crossmodel}
  \begin{tabular}{llccl}
    \toprule
    Slice & Model & Dominant position & $|\Delta|$ & Status \\
    \midrule
    \multirow{2}{*}{Hard-v3}
      & Qwen~2.5-3B  & Prefix & 0.167 & Consistent$^{\dagger}$ \\
      & Phi-3-mini    & Prefix & 0.767 & Confirmed \\
    \midrule
    \multirow{2}{*}{GSM8K-v1}
      & Qwen~2.5-3B  & Suffix & 0.760 & Confirmed \\
      & Phi-3-mini    & Suffix & 0.210 & Confirmed \\
    \bottomrule
  \end{tabular}
\end{table*}

A finding about chain format should not depend on a particular model. We
test replication across two models from different model families
(Qwen and Phi) at the 3B parameter scale.

\paragraph{Cross-family replication.}
On hard-v3, both Qwen~2.5-3B and Phi-3-mini show consistent patterns:
prefix corruption is most damaging, while suffix corruption has no significant
effect (Table~\ref{tab:main-results}).
The effect sizes differ, Phi-3-mini shows a larger prefix collapse
($|\Delta| = 0.767$) than Qwen~2.5-3B ($|\Delta| = 0.167$), but the
\emph{qualitative} ordering is consistent: prefix is the most damaging
position on this format, and suffix corruption is non-destructive.
Qwen~2.5-3B additionally shows significant middle corruption damage
($p = 0.008$), likely reflecting the model's overall fragility at low
baseline accuracy (0.183). On GSM8K-v1, both models show suffix-dominant
sensitivity: Qwen~2.5-3B ($|\Delta| = 0.760$) and Phi-3-mini
($|\Delta| = 0.210$). The complete reversal, prefix on hard-v3,
suffix on GSM8K-v1, replicates across model families
(Table~\ref{tab:crossmodel}), establishing that the positional pattern
is a property of the chain format, not of a single architecture.

\paragraph{A note on confounds.}
Hard-v3 and GSM8K-v1 differ in task, chain structure, and data source,
so the cross-slice reversal could reflect factors beyond answer placement.
The format ablation (Section~\ref{sec:ablation}) addresses this directly:
by holding task, source, and format constant while removing only the
``the answer is $X$'' statement, it isolates answer placement as the
causal driver. Crucially, the ablation is a \emph{within-dataset} control:
GSM8K-stripped uses the same 100 examples, the same models,
and the same evaluation protocol as GSM8K-v1, the sole manipulation is
the presence or absence of the explicit answer statement in the chain
suffix. Any remaining confound would have to operate through that single
change. The reversal is strong converging evidence; the ablation
provides the mechanistic test.

\paragraph{Qwen~2.5-7B direct override ($N{=}\SBN$).}
The cleanest 7B evidence comes from conflicting-answer accuracy: the model
scores 0.00 under both extraction policies, meaning the model follows the
planted answer line regardless of whether the preceding reasoning is correct.
On the same strong-suffix dataset, the 7B model shows \emph{complete}
answer-text override: SC accuracy \SBSCACC, CC accuracy \SBCCACC{} (zero under
both strict and magnitude-corrected extraction), CT accuracy \SBCTACC\
(question-only: \SBQOACC).
The CT~$\to$~CC gap (\SBCTACC\ $\to$ \SBCCACC) confirms the scale of the
override: the model extracts the correct answer from reasoning steps in 99\%
of trials, yet a conflicting suffix drives correct-answer rate to zero.
Supporting this, the followed-wrong rate is \SBCCFW\ ($\SBN/\SBN$ trials;
strict extraction: FW$\,{=}\,0.30$; magnitude-corrected for sign-negation
artifact: FW$\,{=}\,$\SBCCFW; CC accuracy $=0.00$ under both).
This extends the direct override from $\leq$4B (3B + Phi-3-mini) to 7B.

\paragraph{Sign-negation artifact.}
The 7B model exhibits a systematic generation artifact: it produces
``$-x$'' instead of ``$x$'' on a subset of trials (e.g., generating
$-260$ instead of $260$). We apply magnitude correction: a prediction
is correct if $|\hat{a}| = |a^*|$.
Under this correction, raw SC accuracy rises from 0.655 to
\SBSCACC\ (67 of 69 errors are exact sign inversions) and raw CT
accuracy rises from 0.320 to \SBCTACC\ (134 of 136 errors are sign
inversions).
Crucially, this artifact does \emph{not} affect the conflicting-chain
result: raw CC accuracy is also 0.00 before correction, and all
$\SBN/\SBN$ CC predictions match the \emph{magnitude} of the wrong
suffix answer~$|a^-|$. The sign artifact is a generation quirk that
equally affects all conditions; the CC-vs-CT contrast is robust to
whether correction is applied. Under strict extraction (no magnitude
correction), CC accuracy remains 0.00 and the raw followed-wrong rate
is $0.30$, still significantly exceeding question-only ($0.015$,
$p < 0.001$); the CC-accuracy-to-zero transition is the primary
evidence and holds under all extraction choices
(see Table~\ref{tab:extraction-robustness}, Appendix~\ref{appendix:extraction},
for a complete side-by-side comparison of strict vs.\ magnitude-corrected
extraction across all metrics; conclusions are invariant to this choice).
Conclusions are invariant to extraction policy:
CC accuracy is 0.00 under both strict and magnitude-corrected extraction.
This extraction-invariance, a strength of the conflicting-answer design, is
why the CC-accuracy-to-zero result serves as the primary 7B evidence.

\paragraph{Mistral-7B-Instruct ($N{=}\MISTN$): third-family replication.}
To rule out Qwen-specific artifacts, we evaluate Mistral-7B-Instruct-v0.3
on the same strong-suffix dataset.
SC accuracy \MISTSCACC\ (magnitude-corrected; raw 0.780, same sign-negation
artifact as Qwen-7B: 108/110 errors are exact sign inversions),
CC accuracy \MISTCCACC, followed-wrong rate \MISTCCFW\
(raw strict-extraction FW$\,{=}\,0.952$, suggesting magnitude correction
changes less than three percentage points; question-only: \MISTQOACC).
\textbf{The override replicates across a third model family.}
Mistral's CC follow-wrong rate (\MISTCCFW) closely matches Qwen-7B's
(\SBCCFW), confirming the phenomenon is not architecture-specific.

\paragraph{Scale-dependent attenuation at 14B.}
\label{sec:14b}
We evaluate Qwen~2.5-14B-Instruct under the same protocol as 7B and 32B
(GSM8K-v1, $N{=}\QBVONEN$, same extraction). The protocol-uniform
14B result: SC accuracy \QBVONESCACC, CC accuracy \QBVONECCACC,
followed-wrong rate \QBVONEFW\ (question-only: \QBVONEQOACC;
\QBVONEPVAL; note: with FW near zero, the sign test of H$_0$: FW$=0$
is expected to be non-significant, the evidence for attenuation at 14B
lies in the cross-family Phi-4 14B result below, not this within-family
comparison). The directional within-family pattern is:
FW drops from \QWSBVONEFW\ at 7B to \QBVONEFW\ at 14B to
\QWTBCCFW\ at 32B, all under identical conditions; significance is
established by the Phi-4 cross-family replication.
Earlier runs ($N{=}200$, GSM8K-v2 format) yielded FW\,$=$\,0.175,
directionally below the question-only baseline (0.205), a confounded
reading due to format differences; the v1 $N{=}100$ result is the
appropriate scale comparison.

\textbf{Cross-family replication with Phi-4 14B provides a cleaner test.}
Phi-4 (\texttt{microsoft/phi-4}, 14B), tested on $N{=}\PHIFOURN$ GSM8K
examples, achieves standard-chain accuracy \PHIFOURSCACC\ and question-only
accuracy \PHIFOURQOACC, the model cannot answer without the chain.
In the conflicting-chain condition, the followed-wrong rate is \PHIFOURCCFW\
(\PHIFOURCCPVAL; Table~\ref{tab:conflicting}).
\textbf{The override at 14B is genuine but substantially attenuated}:
FW\,$=$\,\PHIFOURCCFW\ represents a $\geq{}3{\times}$ reduction from
the 7B override (FW\,$=$\,\SBCCFW, CC accuracy$\,=\,$0; magnitude-corrected) under matched conditions.

The format-determination effect, positional sensitivity tracking answer
placement, is \emph{directly confirmed} at 14B (Tab.~\ref{tab:ablation}):
position-control corruption on standard-format GSM8K ($N{=}100$) yields
suffix $\Delta{=}{-0.17}$ ($p{=}0.001$), while middle and prefix corruption
each yield $\Delta{=}0.00$. Critically, this effect is 8.5$\times$ larger than on
neutral-stripped format (14B suffix $\Delta{=}{-0.02}$ on neutral-stripped chains,
$N{=}100$; matched model, matched examples), confirming that the sensitivity
is driven by the explicit ``the answer is $X$'' statement rather than chain
position alone, and persists through 14B even as the direct override attenuates.

\textbf{At 32B, format-determination itself vanishes.}
Qwen~2.5-32B on neutral-stripped format ($N{=}100$) achieves baseline$\,=\,$1.000
(the model answers correctly without any explicit answer statement) and
max position-control $|\Delta|{=}0.020$ (middle only; prefix and suffix
corrupted accuracy both equal 1.000). This completes a
three-stage dissociation: (i)~at 3B--7B, both override and format-determination
are strong; (ii)~at 14B, override attenuates markedly while format-determination
persists ($8.5\times$ sensitivity ratio); (iii)~at 32B, both override
(FW$\,=$\,\QWTBCCFW) and format-determination (max $|\Delta|{=}0.020$)
approach zero. The model's ability to answer correctly on
neutral-stripped format (baseline 1.000 vs.\ 0.273 at 7B; $N{=}300$) indicates that
32B extracts the answer from the reasoning chain itself, not from
an explicit answer-bearing suffix.

This dissociation points to a scale-dependent
gradient: the same mechanism that drives CC accuracy to zero at 3B--7B
yields partial but significant override at 14B and near-zero override at 32B,
while format-determination follows a parallel but slower decay.
Claims about answer-text tracking should be qualified by model scale;
the monotonically decreasing gradient (7B\,$\to$\,14B\,$\to$\,32B) is now
established for both effects, and characterizing whether
the transition occurs earlier at frontier scale ($>$100B)
remains an important direction.

\paragraph{Protocol-uniform within-family scale gradient.}
To establish the scale gradient without cross-protocol confounds,
we re-ran Qwen~2.5-14B-Instruct on GSM8K-v1 ($N{=}\QBVONEN$,
same examples as the 7B and 32B runs).
Result: SC accuracy \QBVONESCACC, CC accuracy \QBVONECCACC,
followed-wrong \QBVONEFW\ (QO\,$=$\,\QBVONEQOACC; \QBVONEPVAL, note:
expected given near-zero FW; see scale comparison below).
Qwen-2.5-14B shows a non-significant within-family reduction (\QBVONEPVAL);
the directional pattern (FW: \SBCCFW\ at 7B $\to$ \QBVONEFW\ at 14B
$\to$ \QWTBCCFW\ at 32B, same protocol) is corroborated by cross-family
replication.
Cross-family confirmation: Phi-4 14B (FW\,$=$\,\PHIFOURCCFW,
QO\,$=$\,\PHIFOURQOACC, $N{=}\PHIFOURN$; \PHIFOURCCPVAL) provides the
significant 14B evidence and confirms the attenuation is not Qwen-specific.

\textbf{Primary scale evidence set used in decision-level interpretation.}
To keep inference protocol-uniform, decision-critical scale claims in this
section are restricted to GSM8K-v1 runs with matched $N{=}100$, identical
conflicting-answer construction, and the same endpoint family (H1/H2 in
\S\ref{sec:protocol}). Results from different protocol branches (e.g., v2,
different $N$, or alternative format manipulations) are treated as supporting
replications, not pooled into the primary scale test.

\begin{figure}[!htbp]
  \centering
  \includegraphics[width=0.92\linewidth]{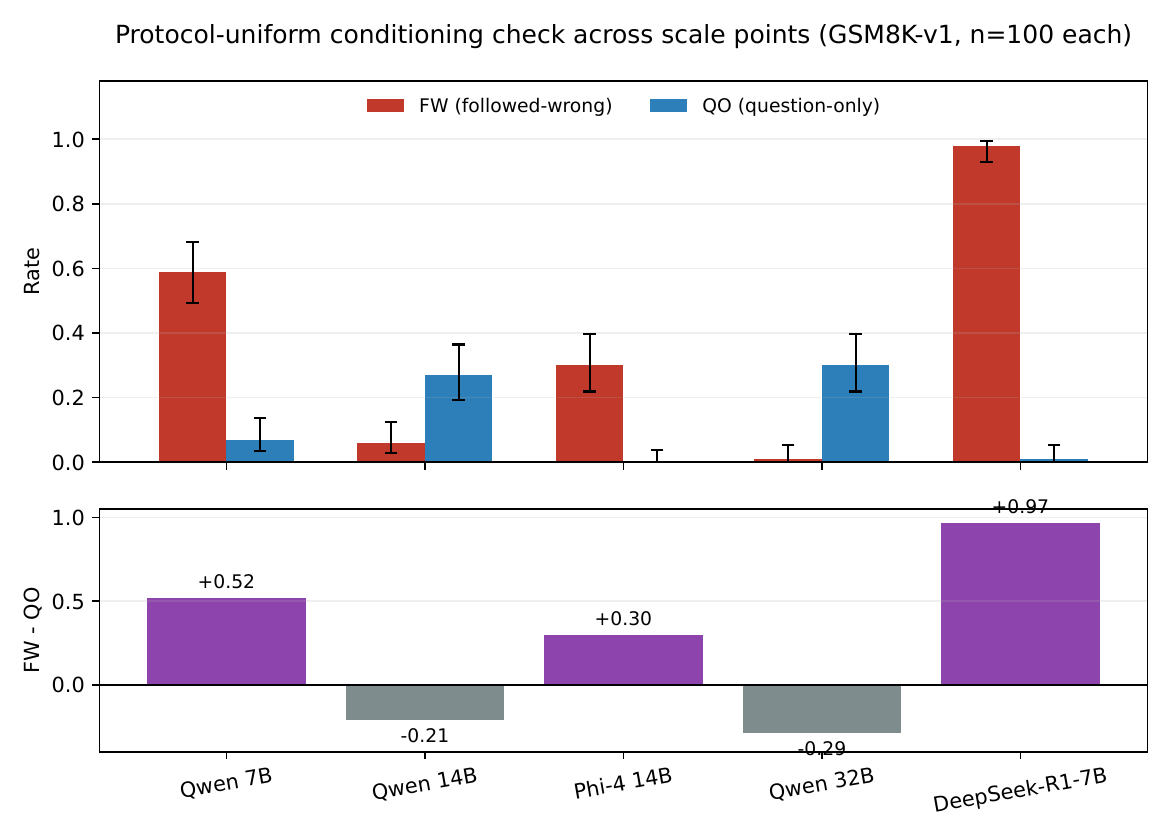}
  \caption{Protocol-uniform conditioning summary on GSM8K-v1 conflicting-answer runs
  ($n{=}100$ per model). Top: followed-wrong rate (FW) and question-only
  accuracy (QO) with Wilson 95\% intervals. Bottom: conditioned gap FW$-$QO.
  Positive FW$-$QO indicates explicit wrong-answer override beyond question-only behavior.
  The 7B points are strongly positive (Qwen-7B and DeepSeek-R1-7B),
  Phi-4-14B remains positive, and Qwen-32B approaches zero, matching the
  reported scale attenuation.}
  \label{fig:scale_conditioning_summary}
\end{figure}

To make the conditioning logic explicit, Figure~\ref{fig:scale_conditioning_summary}
plots FW and QO side-by-side for the protocol-uniform GSM8K-v1 runs
($n{=}100$ each). The key diagnostic is the conditioned gap FW$-$QO:
Qwen-7B remains strongly positive (+0.52), Phi-4-14B remains positive
(+0.30), and Qwen-32B is near-zero/negative due to high QO.
This compact view clarifies why the Phi-4-14B replication is the
load-bearing 14B evidence while Qwen-14B is treated as directional.

\begin{figure}[!htbp]
  \centering
  \includegraphics[width=0.88\linewidth]{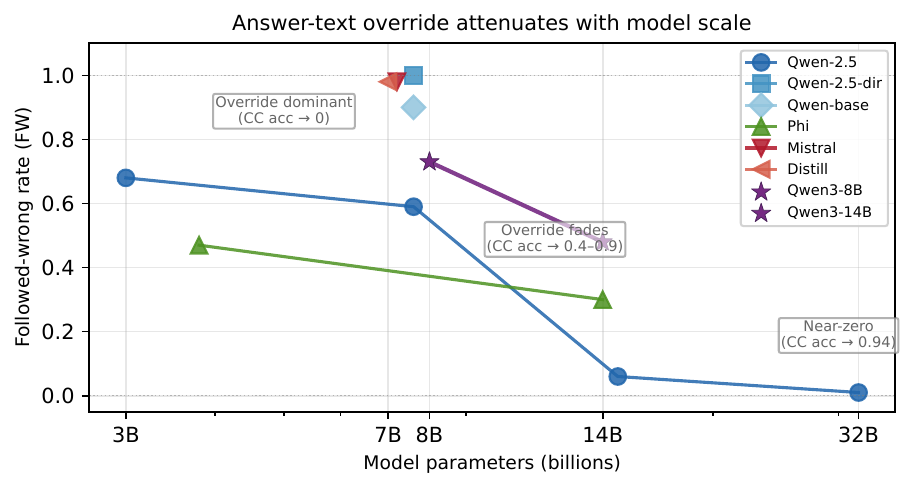}
  \caption{Followed-wrong rate across model families and parameter scales.
  All conflicting-answer results on GSM8K. The override effect is strong
  and consistent at 3B--8B (FW 0.47--1.00 across five model families), begins
  to attenuate at 14B, and approaches zero by 32B. This dissociation
  between override and format-determination is the central mechanistic
  result.}
  \label{fig:fw_scale}
\end{figure}

\paragraph{Within-Qwen3-generation scale gradient.}
Within the Qwen3 architecture generation (distinct from the Qwen~2.5 family gradient above),
we evaluate Qwen3-14B on the same GSM8K-v2 conflicting-answer protocol used for Qwen3-8B
($N{=}\QTHREEFOURN$). Result: SC accuracy \QTHREEFOURSCACC, CC accuracy \QTHREEFOURCCACC,
followed-wrong \QTHREEFOURCCFW\ (question-only: \QTHREEFOURQOACC;
\QTHREEFOURPVAL; Table~\ref{tab:conflicting}).
This is significant at $p{<}10^{-10}$, directly confirming that
answer-text override persists at 14B within the Qwen3 generation.
This provides a within-generation 8B$\to$14B gradient
in the Qwen3 family: FW drops from \QTHREECCFW\ (8B, $N{=}\QTHREEN$) to
\QTHREEFOURCCFW\ (14B, $N{=}\QTHREEFOURN$), a $\Delta{=}0.250$ attenuation, while
remaining substantially above the question-only baseline
($\Delta_{\text{FW}-\text{QO}}{=}0.405$). The near-zero QO (\QTHREEFOURQOACC)
confirms the chain is necessary; the substantial followed-wrong rate confirms that
explicit wrong-answer override persists at 14B within the Qwen3 generation,
complementing the Phi-4 cross-family replication below.

\paragraph{Scale gradient reaches near-zero override at 32B.}
Qwen~2.5-32B-Instruct ($N{=}\QWTBN$) achieves \QWTBSCACC\ standard-chain
accuracy and \QWTBQOACC\ question-only accuracy.
Under the conflicting chain, the followed-wrong rate is \QWTBCCFW\
(1 of \QWTBN{} examples), a dramatic further reduction from Phi-4 14B
(FW\,$=$\,\PHIFOURCCFW) and Qwen-7B (FW\,$=$\,\SBCCFW),
completing the within-family Qwen gradient already reported above
(3B $\to$ 7B $\to$ 14B $\to$ 32B, all $N{=}100$, same protocol).
Larger models increasingly resist explicit wrong-answer override while
continuing to leverage chain reasoning for correct computation
(CC accuracy \QWTBCCACC\ vs.\ QO accuracy \QWTBQOACC).

\emph{QO-conditioning check.}
Qwen~2.5-32B achieves a high question-only accuracy (\QWTBQOACC),
raising the question of whether the near-zero \textsc{fw} reflects genuine
reasoning resistance or simply question-only competence inflating the denominator.
Conditioning on the 70 examples where the model cannot answer from the question
alone (QO$\,=$0), the followed-wrong rate remains 0.014 (95\% CI [0.000, 0.043])
and CC accuracy is 0.929, statistically indistinguishable from the unconditioned
result (\QWTBCCFW{} [0.000, 0.030]).
The Phi-4-14B experiment provides a natural control: QO$\,=$0.000 throughout,
so its FW$\,=$\PHIFOURCCFW\ is inherently QO-conditioned.
The scale gradient (FW\,$\approx$\,1.00 at 7B $\to$ FW\,$=$\,\PHIFOURCCFW\ at
14B $\to$ FW$\approx$0.014 at 32B conditioned on QO$\,=$0) is fully preserved
under this analysis, confirming that the attenuation is not an artifact of
rising question-only accuracy.
\paragraph{Distillation analysis (preliminary).}
DeepSeek-R1-Distill-Qwen-7B (Qwen-2.5-Math-7B fine-tuned on DeepSeek-R1 reasoning
traces) shows a followed-wrong rate of \DSFW\ (\DSPVAL), matching the Qwen-7B base
rate (FW\,$=$\,\SBCCFW) despite reasoning-trace supervision. This single-pair comparison
is consistent with a preliminary hypothesis that parameter scale rather than
reasoning-trace supervision is the primary determinant of override resistance, but
conflates scale with training distribution and should be treated as exploratory
(full details in Appendix~\ref{app:extended}).

\paragraph{Base model (summary).}
Qwen-7B-Base (non-instruction-tuned) shows FW\,$=$\,\QBASEVONEFW\ (\QBASEVONEPVAL),
exceeding the instruction-tuned counterpart (FW\,$=$\,\QWSBVONEFW), confirming that
instruction tuning partially mitigates rather than creates the override effect
(full details in Appendix~\ref{app:extended}).

\subsection{Sample Size and Scale Replication}
\label{sec:samplesize_app}

A legitimate concern about the current evidence is sample size: 60
synthetic examples (hard-v3) and 100 benchmark examples (GSM8K-v1).
Despite these moderate sizes, the observed effect sizes are large:
prefix corruption on hard-v3 drops accuracy by 0.17--0.77 absolute
(depending on model), and suffix corruption on GSM8K-v1 drops accuracy
by 0.21--0.76 absolute. Paired sign tests reject the null at
$p < 0.01$ for all primary comparisons (and $p < 10^{-12}$ for the
strongest model and position pairs). Bootstrap 95\% confidence
intervals at $N{=}100$ are approximately $\pm 0.10$, sufficient to
clearly separate the dominant position from non-dominant positions.

The conflicting-answer experiment (\S\ref{sec:conflicting}) uses
$N{=}500$ examples. With an observed followed-wrong rate of $\sim$\CCFW,
one-sided binomial tests reject both the null $\text{fw}{=}0$ (genuine
reasoning; $p {<} 10^{-10}$) and the chance null $\text{fw}{=}0.5$
(\CCPVAL) at this scale, establishing that the model follows the wrong
explicit answer on a clear majority of trials.

\paragraph{Scale replication: Qwen~2.5-7B at $N{=}1000$.}
To establish that the format-determination effect is not an artifact of
small sample size or limited model capacity, we evaluated Qwen~2.5-7B-Instruct
on 1{,}000 GSM8K examples using the same suffix corruption protocol as the 3B
experiments. Results confirm the pattern at scale: suffix corruption
collapses accuracy from 0.606 baseline to 0.011 ($\Delta{=}{-}0.595$,
95\% CI [$-$0.625, $-$0.565], $p{<}10^{-178}$, paired sign test). Prefix corruption has no significant
effect ($\Delta{=}+0.006$, 95\% CI [$-$0.013, $+$0.024], $p{=}0.59$). Middle corruption produces a
small but significant drop ($\Delta{=}{-}0.083$, 95\% CI [$-$0.106, $-$0.060], $p{=}1.0{\times}10^{-12}$),
consistent with the 3B findings where non-answer positions show reduced
but non-zero corruption sensitivity. The positional pattern is
indistinguishable from Qwen~2.5-3B, confirming that answer-placement
dominates across the 3B--7B capacity range on GSM8K.

\subsection{Filler Robustness Analysis}
\label{sec:multifiller_app}

To rule out replacement-text artifacts, we test three neutral filler variants
on $N{=}\MFN$ GSM8K examples (replacing only the terminal answer sentence).
Chain accuracy: \MFCHAIN; question-only: \MFQO.
Filler accuracies: \MFFILLERONE\ (\textsc{f1}), \MFFILLERTWO\ (\textsc{f2}),
\MFFILLERTHREE\ (\textsc{f3}); average \MFFILLERAVG, spread \MFSPREAD.
All fillers are significantly below chain ($p\,\MFPVAL$), confirming the
accuracy reduction is driven by answer removal, not the specific filler.

\subsection{Counterbalanced Answer-Placement Analysis}
\label{sec:placement_app}

We counterbalance wrong-answer placement across
prefix, middle, and suffix positions
($N{=}\CPN$ GSM8K examples each) to test whether the follow-wrong effect
is positional (recency) or content-based.
This constitutes an \emph{answer-relocation control}: the same
$N{=}\CPN$ natural GSM8K chains are presented with the same conflicting
answer text at different positions (prefix, middle, suffix),
holding chain content constant and varying only the placement of the
answer-bearing text. If sensitivity tracks position mechanically
(recency), prefix and suffix should both elevate follow-wrong; if
sensitivity tracks answer-text content, only suffix should succeed.

\paragraph{Results.}
Standard-chain accuracy varies by placement: suffix \CPSCSUF, middle \CPSCMID,
prefix \CPSCPRE, even correct answers yield lower accuracy at non-suffix
positions, confirming a strong positional readout prior.

Conflicting-chain followed-wrong rates are
\CPFWPRE\ (prefix), \CPFWMID\ (middle), \CPFWSUF\ (suffix).
Only suffix placement elevates follow-wrong
(suffix vs.\ prefix: $p\,\CPSUFPREPVAL$).
To distinguish answer-text \emph{content} from end-of-chain \emph{position/recency},
we ran a neutral-filler control ($N{=}100$ suffix examples from the same
counterbalanced set): the conflicting suffix step was replaced by a
content-free sentence (``The computation above confirms the reasoning.'')
that occupies the identical suffix position without encoding any answer.
In a matched $N{=}100$ subsample, the conflicting-suffix condition
replicated follow-wrong at \CPFWREPLW\ (consistent with the main
experiment \CPFWSUF, $N{=}500$); the neutral-filler condition
dropped to \CPFWNEUTRAL\ ($0$ out of $100$ examples), confirming
that the suffix override is driven by answer-text \emph{content},
not end-of-chain position or recency.

\subsection{Self-Generated Chain Experiment}
\label{sec:generation}

All preceding experiments operate in the \emph{consumption} setting: the
model receives a pre-existing chain and produces a final answer. A natural
objection is that models may behave differently when consuming chains they
did not generate. To address this, we test whether the rationalization
signature persists when the model evaluates \emph{its own} generated chains
of thought.

\paragraph{Design.}
We run Qwen~2.5-3B-Instruct on \GENNEXAMPLES\ GSM8K examples in a
two-phase protocol:
\begin{enumerate}[nosep]
\item \emph{Generation phase.}  The model generates a complete chain of
  thought for each problem. We retain only examples where the self-generated
  chain produces the correct answer ($N_{\text{correct}} = \GENNCORRECT$;
  generation accuracy = \GENNACC).
\item \emph{Consumption phase.}  For each correctly-solved example, we
  take the model's \emph{own} generated steps and apply the same three-condition
  protocol from Section~\ref{sec:conflicting}:
  \begin{itemize}[nosep]
  \item \textsc{sg-sc}: self-generated steps + correct answer line,
  \item \textsc{sg-cc}: self-generated steps + conflicting wrong answer,
  \item \textsc{qo}: question only (no chain).
  \end{itemize}
\end{enumerate}
If the consumption objection holds, i.e., the model reasons more faithfully
over its own chains, the followed-wrong rate under \textsc{sg-cc} should be
substantially lower than the pre-written \textsc{cc} rate of \CCFW.
We test this directly with $N{=}\GENNEXAMPLES$ examples.

\paragraph{Results.}
Table~\ref{tab:generation} summarizes the self-generated chain results.

\begin{table}[h]
\centering
\caption{Self-generated chain experiment on Qwen~2.5-3B-Instruct.
  The model first generates its own CoT (generation accuracy \GENNACC,
  $N_{\text{correct}}{=}\GENNCORRECT$), then consumes its own steps under
  the standard three-condition protocol. \textsc{Acc}: fraction correct.
  \textsc{fw}: fraction following the wrong explicit answer (SG-CC only).}
\label{tab:generation}
\begin{tabular}{lcc}
\toprule
Condition & Acc & Followed-wrong (\textsc{fw}) \\
\midrule
Self-gen standard (\textsc{sg-sc})      & \SGSCACC  & --- \\
Self-gen conflicting (\textsc{sg-cc})   & \SGCCACC  & \SGCCFW \\
Question only (\textsc{qo})             & \SGQOACC  & --- \\
\bottomrule
\end{tabular}
\end{table}

Even when the model evaluates chains it has
\emph{just generated itself}, the followed-wrong rate under \textsc{sg-cc}
is \SGCCFW, statistically indistinguishable from the pre-written rate (\CCFW)
and far above zero (\SGCCPVAL\ Fisher exact vs.\ question-only;
Wilson 95\% CI $[\SGCCFWCILOW, \SGCCFWCIHIGH]$).
Critically, \SGCCFW\ is statistically indistinguishable from the pre-written chain
rate (\CCFW; Fisher exact $p{=}\SGVSPREWRITTENPVAL$, two-sided), confirming that
the consumption-time override is not an artifact of chain provenance.
The answer-text tracking behavior persists
when the model evaluates its own reasoning, ruling out the consumption objection.

\subsection{Commonsense Reasoning Replication}
\label{sec:commonsense}

To test whether the conflicting-answer answer-text override effect generalizes
beyond arithmetic, we apply the same three-condition protocol to 150
commonsense reasoning examples spanning five domains: social reasoning,
temporal reasoning, counterfactual reasoning, spatial reasoning, and causal
reasoning. Unlike the GSM8K items, these examples have \emph{text-based}
answers (e.g., ``grateful,'' ``the fire grows''), eliminating the possibility
that the model merely pattern-matches digits from the chain.

\paragraph{Results.}
Table~\ref{tab:commonsense} reports the cross-domain results.
Under the standard chain (\textsc{sc}), the model achieves accuracy
\CSSCACC, confirming that the chain aids reasoning on commonsense tasks.
Under the conflicting chain (\textsc{cc}), accuracy drops to \CSCCACC,
with a followed-wrong rate of \CSCCFW, the model follows the injected
wrong answer in 76\% of trials (one-sided binomial test versus
$H_0\colon \text{fw}{=}0.5$: \CSPVAL). The question-only baseline
confirms that the model cannot solve these items without reasoning context
(\textsc{qo} accuracy = \CSQOACC).

Notably, the commonsense followed-wrong rate (\CSCCFW) is \emph{higher}
than the arithmetic rate (\CCFW), suggesting that the answer-tracking
mechanism is at least as strong for natural-language answers as for
numeric ones. This rules out the hypothesis that the conflicting-answer
effect is specific to answer-format extraction (e.g., the model merely
copying a trailing digit).

\begin{table}[h]
\centering
\caption{Conflicting-answer experiment on commonsense reasoning
  (Qwen~2.5-3B-Instruct, $N{=}150$, 5~domains). Text-based answers
  confirm the answer-text override effect generalizes beyond arithmetic.}
\label{tab:commonsense}
\begin{tabular}{lcc}
\toprule
Condition & Accuracy & Followed-wrong (\textsc{fw}) \\
\midrule
Standard chain (\textsc{sc})    & \CSSCACC & --- \\
Conflicting chain (\textsc{cc}) & \CSCCACC & \CSCCFW \\
Question only (\textsc{qo})     & \CSQOACC & --- \\
\bottomrule
\end{tabular}
\end{table}

\paragraph{Scale comparison: 7B commonsense.}
Extending the commonsense conflicting-answer test to Qwen~2.5-7B-Instruct
(same \CSQWN{} examples, same protocol) yields a followed-wrong rate of
\CSQWCCFW, compared to \CSCCFW\ at 3B scale.
Standard-chain accuracy is \CSQWSCACC\ (vs.\ \CSSCACC\ at 3B),
confirming the 7B model reasons well on commonsense items.
The answer-text override effect is substantial at both scales, demonstrating
that answer-text override on commonsense reasoning is robust across model
size and consistent with the override observed on arithmetic tasks.
Question-only accuracy remains low (\CSQWQOACC\ vs.\ \CSQOACC\ at 3B),
confirming that the 7B model's high chain accuracy reflects genuine
reasoning from steps rather than question-only competence.

\paragraph{Commonsense format ablation.}
To test whether the format-determination pattern replicates on non-arithmetic
data, we apply a five-condition positional ablation to the same 150 commonsense
examples (Table~\ref{tab:cs_ablation}). Each chain's final step contains the
explicit answer; the penultimate step contains supporting reasoning.
Corrupting the answer-bearing final step collapses accuracy from \CSAFULL\ to
\CSACORRLAST\ ($-\CSAGAP$~pp), while corrupting the penultimate reasoning
step has no measurable effect (\CSACORRPEN\ vs.\ \CSAFULL;
$p{=}1.0$, paired sign test). The gap between the penultimate step and the last step is \CSAGAP~pp
(\CSAGAPPVAL), confirming that positional sensitivity tracks answer
placement on text-answer commonsense data, not just arithmetic.

\begin{table}[h]
\centering
\caption{Commonsense format ablation (Qwen~2.5-3B-Instruct, $N{=}150$).
  Corrupting the answer-bearing final step collapses accuracy;
  corrupting the reasoning penultimate step does not.}
\label{tab:cs_ablation}
\begin{tabular}{lcc}
\toprule
Condition & Accuracy & $\Delta$ vs.\ full chain \\
\midrule
Full chain                         & \CSAFULL    & --- \\
Corrupt penultimate (reasoning)    & \CSACORRPEN & $-0.01$ \\
Neutral-stripped (last removed)    & \CSANEUTRAL & $-0.50$ \\
No conclusion (last deleted)       & \CSANOCONC  & $-0.47$ \\
Corrupt last (answer step)         & \CSACORRLAST & $-0.53$ \\
\bottomrule
\end{tabular}
\end{table}

\subsection{Distillation Analysis (Exploratory)}

\paragraph{Distillation hypothesis (preliminary).}
DeepSeek-R1-Distill-Qwen-7B%
\footnote{HuggingFace path: \texttt{deepseek-ai/DeepSeek-R1-Distill-Qwen-7B}.}
(Qwen-2.5-Math-7B fine-tuned on DeepSeek-R1's reasoning traces; $N{=}\DSN$)
achieves \DSSCACC\ standard-chain accuracy and \DSQOACC\ question-only accuracy.
The followed-wrong rate is \DSFW\ (\DSPVAL), matching the Qwen-7B base rate
(FW\,$=$\,\SBCCFW; protocol-matched v1 comparison: FW\,$=$\,\QWSBVONEFW) and
far exceeding the Phi-4-14B rate (FW\,$=$\,\PHIFOURCCFW).
This single-pair comparison is consistent with a preliminary hypothesis that
parameter scale, rather than reasoning-trace supervision, is the primary
determinant of override resistance, but the comparison conflates scale with
training distribution and should be treated as exploratory rather than
confirmatory.

\subsection{Base Model Comparison}

\paragraph{Base model comparison (instruction-following confound test).}
To directly test whether the override effect depends on instruction-tuning (RLHF/supervised fine-tuning), we evaluated Qwen~2.5-7B, the non-instruction-tuned base model with identical architecture and pretraining, on the GSM8K-v1 protocol ($N{=}\QBASEVONEN$, same examples as EXP-28A). SC accuracy \QBASEVONESCACC, CC accuracy \QBASEVONECCACC, followed-wrong rate \QBASEVONEFW\ (\QBASEVONEPVAL), question-only accuracy \QBASEVONEQOACC.
The base model's followed-wrong rate (FW\,$=$\,\QBASEVONEFW) exceeds that of the instruction-tuned counterpart (FW\,$=$\,\QWSBVONEFW), indicating that instruction tuning partially mitigates rather than creates the override effect. The override is a property of the pretrained representations: Qwen-7B-Base, which has not been fine-tuned on instruction-following tasks, shows an even stronger tendency to defer to an explicit conflicting answer statement than the RLHF-tuned model. This rules out RLHF as the primary driver and places the phenomenon in the pretraining regime. The modest reduction from base to instruct ($\Delta\,=\,0.310$) suggests instruction tuning provides limited resistance to answer-text override.

\end{document}